\documentclass[sigconf, nonacm]{acmart}  

\usepackage{booktabs}
\usepackage{xcolor}
\usepackage{hyperref}
\usepackage{xspace}

\usepackage{enumitem}
\usepackage{caption}
\usepackage{cleveref}
\usepackage{multirow}
 
\usepackage{amsmath}
\usepackage{makecell}
\usepackage{subcaption}
\usepackage{graphicx}
\usepackage{amsfonts}
\usepackage{algorithm}
\usepackage{algorithmic}
\usepackage[normalem]{ulem}
\usepackage{colortbl}

\newcommand{\tofill}[1]{\cred{$\times\times\times$}\xspace}

\newcommand{\cred}{\textcolor{red}}

\newcommand{\sys}{\textsc{DiTango}\xspace}
\newcommand{\sysurl}{\url{https://github.com/thu-pacman/ChituDiffusion}}

\AtBeginDocument{%
  }



\begin{document}

\title{\sys: Cost-Effective Parallel Diffusion Generation with Selective Attention State Reuse}

\author{Yuyang Chen}
\email{chen-yy20@sjtu.edu.cn}
\affiliation{%
  \institution{Shanghai Jiao Tong University, Shanghai AI Laboratory}
  \city{Shanghai}
  \country{China}
}

\author{Runxin Zhong}
\email{zhongrx24@mails.tsinghua.edu.cn}
\affiliation{%
  \institution{Tsinghua University}
  \city{Beijing}
  \country{China}
}

\author{Zan Zong}
\email{zongzan@ustb.edu.cn}
\affiliation{%
  \institution{University of Science and Technology Beijing}
  \city{Beijing}
  \country{China}
}

\author{Hengjie Li}
\email{lihengjie@pjlab.org.cn}
\affiliation{%
  \institution{Shanghai AI Laboratory}
  \city{Shanghai}
  \country{China}
}

\author{Yuyang Jin}
\email{jinyuyang@tsinghua.edu.cn}
\affiliation{%
  \institution{Tsinghua University}
  \city{Beijing}
  \country{China}
}

\author{Jidong Zhai}
\email{zhaijidong@tsinghua.edu.cn}
\affiliation{%
  \institution{Tsinghua University}
  \city{Beijing}
  \country{China}
}

\begin{abstract}

Recent advances in AI-generated content have driven widespread adoption of Diffusion Transformers (DiTs) for high-resolution, long-duration content generation. While parallelization techniques accelerate diffusion inference, they face significant scalability challenges due to excessive communication overhead in multi-node environments.

We observe that sequence partitions in Context Parallelism (CP) exhibit distinct heterogeneity: spatially proximate partitions contribute more significantly to attention computation results. By mapping this heterogeneous pattern to hierarchical communication topology, we can access high-contribution partitions with reduced communication cost. This insight motivates our novel selective attention state mechanism that strategically balances partial attention computation and historical result reuse across denoising steps.

We present \sys{}\footnote{This is a preprint of the paper accepted for publication at HPDC '26. 
The final authenticated version will be available at \href{https://doi.org/10.1145/3806645.3807581}{10.1145/3806645.3807581}. Code is available at \sysurl.}, an efficient parallel framework for DiT generation. \sys features an anchor-guided state selection planner that optimizes computation-reuse decisions for each partition, complemented by a runtime that orchestrates efficient state-centric operations. This design achieves superior system efficiency while preserving generation quality.

Experimental evaluation on popular diffusion models demonstrates that \sys achieves up to $1.9\times$ end-to-end and $3.2\times$ attention speedup with near-linear scaling in multi-node settings, while maintaining generation quality comparable to state-of-the-art approaches.

\end{abstract}

\maketitle

\vspace{20px}
\section{Introduction}

Diffusion models have become a foundational paradigm for AI-generated content (AIGC), underpinning state-of-the-art image and video synthesis systems. Driven by the trend toward long-sequence generation (e.g., longer videos with higher spatial and temporal resolution), Diffusion Transformers (DiTs)~\cite{dit} have emerged as a dominant backbone, enabling rapid progress in both closed-source models (e.g., Sora~\cite{sora}, Kling~\cite{kling}) and open-source models (e.g., Wan~2.2~\cite{wan}, HunyuanVideo~\cite{hunyuan}, CogVideoX~\cite{cogvideox}).


Despite the impressive progress in generation quality, DiT-based generators remain prohibitively slow at inference time. For example, Wan2.2~\cite{wan} takes over one hour on a single NVIDIA A100 GPU to generate a 5-second 720p video.
This latency is largely dominated by the full attention computation in DiT architectures: unlike autoregressive LLMs that amortize attention costs through KV caching, DiT must compute full attention over the entire spatiotemporal sequence in each denoising step.

\begin{figure}[t]
    \centering
    \includegraphics[width=1\linewidth]{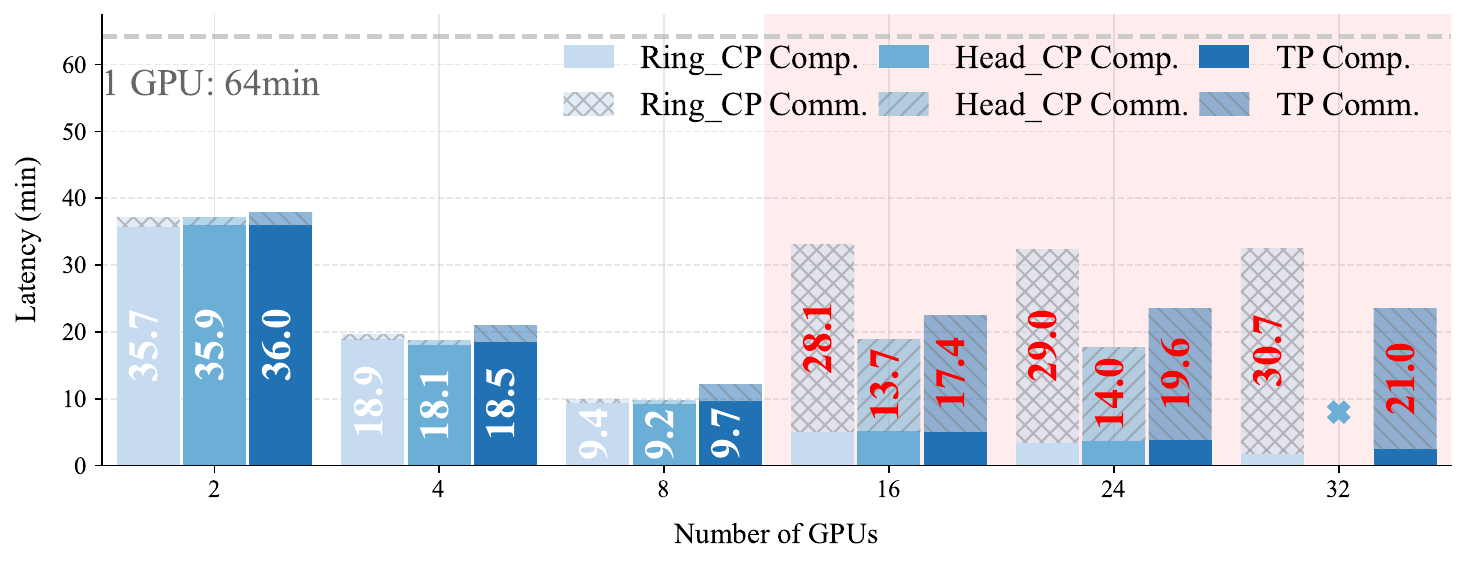}
    \vspace{-1em}
    \caption{Parallel inference performance breakdown for multi-GPU deployments with 8 GPUs per node.}
    \vspace{-1em}
    \label{fig:intro-attn-breakdown}
\end{figure}

\begin{figure}[t]
    \centering
    \includegraphics[width=1\linewidth]{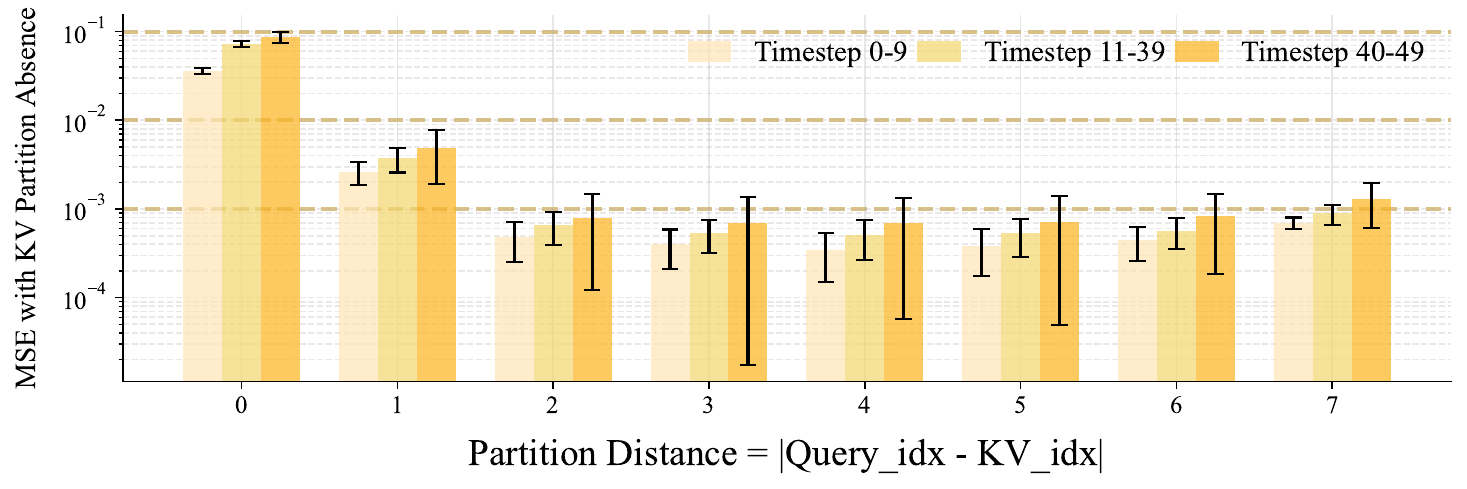}
    \caption{Attention computation contribution is highly unequal with distance-based decay across different CP partitions.}
    \label{fig:contribution}
\end{figure}  

\begin{figure*}[!ht]
    \centering
    \includegraphics[width=0.9\textwidth]
    {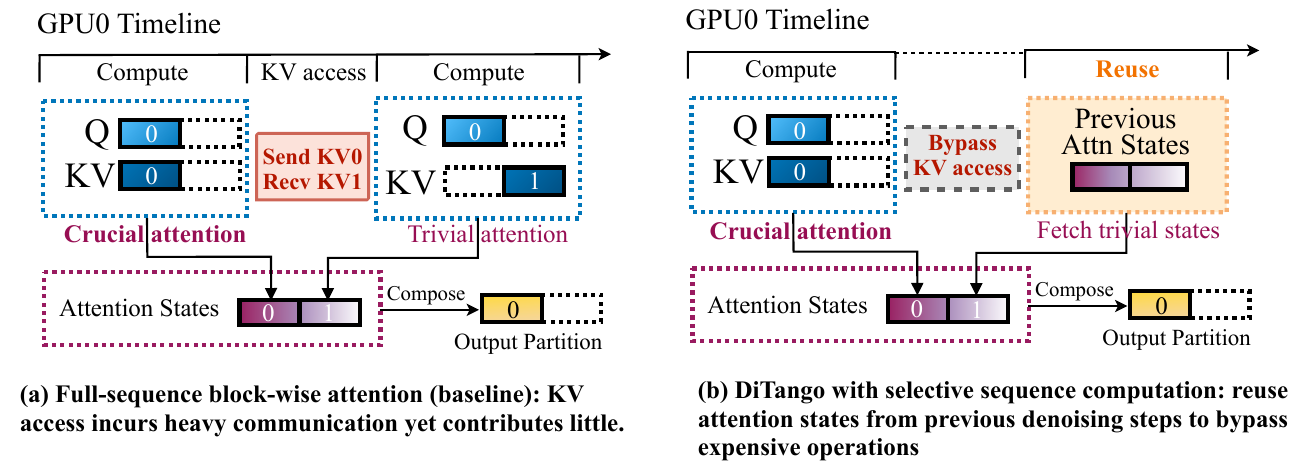}
    \caption{Comparison between full-sequence attention and \sys with selective computation via state reuse.}
    \label{fig:sys_comparision}
\end{figure*}

From a systems perspective, parallel inference offers a promising approach to accelerate DiT-based diffusion generation. However, Data Parallelism (DP) and Pipeline Parallelism (PP) provide limited benefits due to the inherent single-sample nature of diffusion generation (batch size $=1$). Consequently, practical deployments primarily leverage two strategies: Tensor Parallelism (TP)\cite{megatronlm} and Context Parallelism (CP)\cite{ringattention,ulysses,usp}.
TP shards model weights across GPUs and employs collective operations to assemble intermediate results, while CP\cite{ulysses,ringattention,usp} partitions tokens along the sequence dimension with communication for remote Key-Value partition access during attention. When attention computation dominates the runtime, these data transfers can be effectively overlapped with independent computation.

However, these parallel inference solutions still fall short of practical requirements. Our profiling of Wan-14B\cite{wan} generating a 81-frame 720p video (75,600 tokens) on NVIDIA H20s reveals that even with 8 GPUs in a single node, generation takes around 10 minutes. More critically, scaling to multiple nodes often degrades performance due to communication bottlenecks, as shown in Fig.~\ref{fig:intro-attn-breakdown}.

This performance degradation stems from the communication-intensive nature of both TP and CP. TP requires frequent collective operations for sharded model weights, while CP variants either need All-to-All exchanges (Head-CP) or multiple rounds of P2P transfers (Ring-CP) to access global Key-Value sequence. In multi-node settings, limited inter-node bandwidth severely constrains both collective and ring communication patterns. Furthermore, the compute-communication overlap that proves effective within a single node becomes significantly less efficient across nodes, exacerbating the end-to-end performance impact.

To address the performance bottleneck in parallel inference, we identify a key pattern in CP-partitioned sequences for Diffusion inference: \emph{attention contributions from different partitions exhibit strong spatial locality}. 
As shown in Fig.~\ref{fig:contribution}, we conduct a systematic analysis on different models and prompts under CP size=8 by measuring the Mean Squared Error (MSE) of attention outputs when excluding specific KV partitions. Our findings reveal that partitions contribute highly unequally to the final attention results. For instance, excluding the nearest partition (distance$=0$) yields MSE of $10^{-2}$, while distant partitions (distance$\geq2$) contribute merely $10^{-4}$—a difference of 2-3 orders of magnitude. Attention contributions systematically decay with spatial distance that KV Partitions closer to the query (smaller $|\text{Q\_idx} - \text{KV\_idx}|$) consistently dominate the computation, demonstrating clear spatial locality. This spatial locality persists across diverse models, different DiT layers and various input prompts---indicating it is an intrinsic property of diffusion attention rather than an artifact of specific configurations.

This observation unveils a compelling optimization opportunity: by aligning locality priorities with the distributed system topology—mapping high-contribution partitions to nodes with efficient intra-node communication while placing low-contribution ones across nodes—we can strategically compute only the most cost-effective partitions and reuse historical results for less critical ones. This contribution-aware approach enables substantial efficiency gains without sacrificing generation quality.

In this paper, we propose \sys, an efficient and scalable generation system in distributed DiT inference.
\sys exploits the heterogeneous contribution in CP-partitioned attention through a selective computation strategy: it prioritizes cost-effective KV pairs with high contribution and low communication costs, while reusing attention states (partial attention computation results) from previous denoising steps for less critical computations.
As the two-partition example illustrated in Fig.\ref{fig:sys_comparision}, the conventional full attention (Fig.\ref{fig:sys_comparision}a) performs costly communication to access remote KV partition in step 1, despite their trivial contribution to the final result. In contrast, \sys (Fig.~\ref{fig:sys_comparision}b) selectively computes crucial partition locally while bypassing expensive communication by reusing attention states from previous denoising steps for trivial contributions. This design elegantly preserves the mathematical properties of attention computation while significantly reducing communication overhead.

However, implementing efficient attention state reuse presents two fundamental challenges:

\begin{enumerate}
\item \textbf{Complex Selection Space} - The reuse decision for each state must balance multiple factors: computational contribution, communication overhead, and reuse error. Making optimal selections across numerous states to achieve both good performance and quality becomes jointly complex.

\item \textbf{Irregular Access Patterns} - Uneven selection disrupts the structured pattern in context parallelism, resulting in interleaved local memory accesses and heterogeneous communications that are inherently difficult to parallelize.
\end{enumerate}

To address these challenges, \sys introduces two key components:

\begin{enumerate}
\item \textbf{Anchor-guided Selection Planner} - By theoretically modeling attention state errors and designing anchor step-guided error prediction, we precisely identify states with minimal impact for reuse to optimize the performance-quality trade-off.
\item \textbf{State-centric Parallel Runtime} - We schedule attention states with similar characteristics in groups and design a series of state-centric manipulations for efficient computation and retrieval. A specialized pipeline orchestrates these operations to maximize parallel computation and communication efficiency.
\end{enumerate}

Our contributions are summarized as follows:
\begin{itemize}
    \item We identify and analyze the spatial locality in attention contributions during DiT inference, revealing an opportunity to selectively reuse attention states based on their contribution significance and communication costs.
    
    \item We design and implement \sys, a high-performance generation system that leverages theoretical error modeling to make precise reuse decisions while maintaining efficient execution through group-wise state management.

    \item Extensive evaluations on Wan2.1~\cite{wan} and HunyaunVideo~\cite{hunyuan} demonstrate that \sys achieves near-linear scalability across 32 GPU cards, delivering up to 1.9× end-to-end speedup and 3.2× core attention speedup while maintaining superior generation quality compared to state-of-the-art frameworks~\cite{xdit,teacache,videosys2024,pab}.

\end{itemize}

\section{Background and Motivation}

\subsection{Background}

\subsubsection{Diffusion Transformers}
Diffusion Transformers (DiTs)\cite{dit} generate content through an iterative denoising process, typically requiring 20-50 steps. In each step, the same transformer model processes latent representations to gradually refine them from random noise to high-quality content. Unlike traditional U-Net architectures, DiTs leverage transformer-based designs that excel at handling long sequences through their attention mechanisms. Recent DiT architectures\cite{cogvideox, mochi, stepvideo, hunyuan, wan} have widely adopted 3D full attention, which unifies spatial and temporal dimensions into a single sequence for processing. While this unified attention approach significantly enhances generation quality and temporal consistency, attention operations dominate computational cost, consuming over 70\% of total generation time across all denoising steps.

\subsubsection{Parallel Methods for Diffusion Generation}
\begin{figure}[!t]
\centering
\subfloat[Tensor Parallelism]{%
\includegraphics[width=\linewidth]{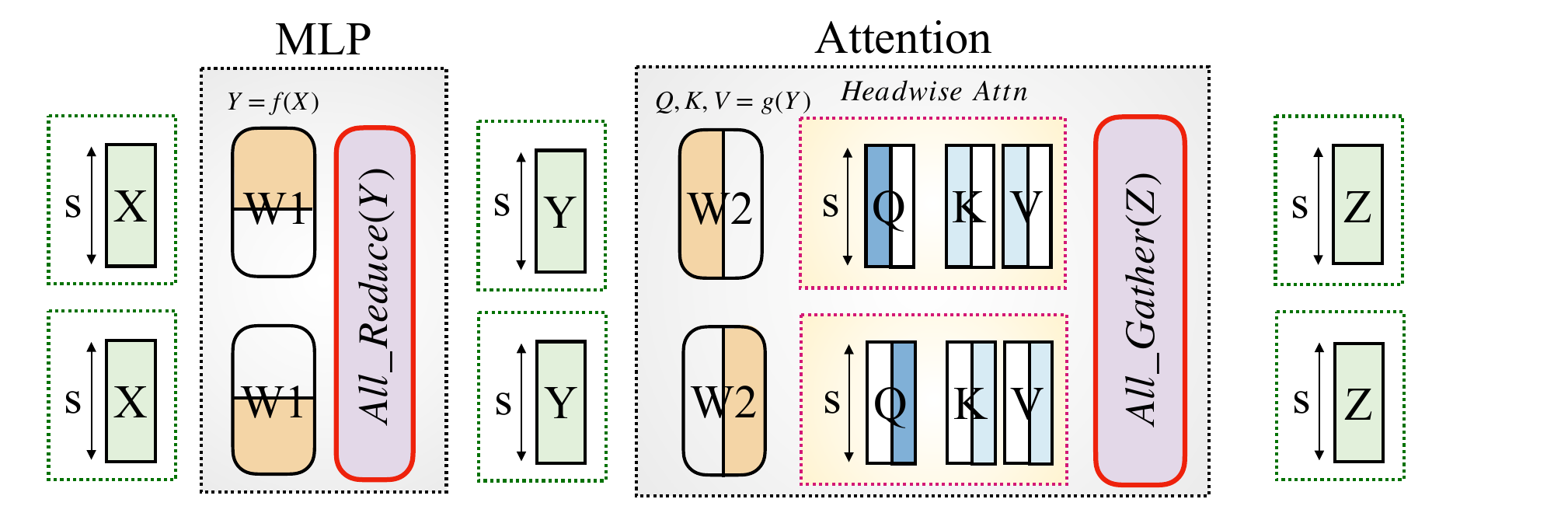}%
\label{fig:tp}%
}\vfill
\subfloat[Context Parallelism with head-wise attention]{%
\includegraphics[width=\linewidth]{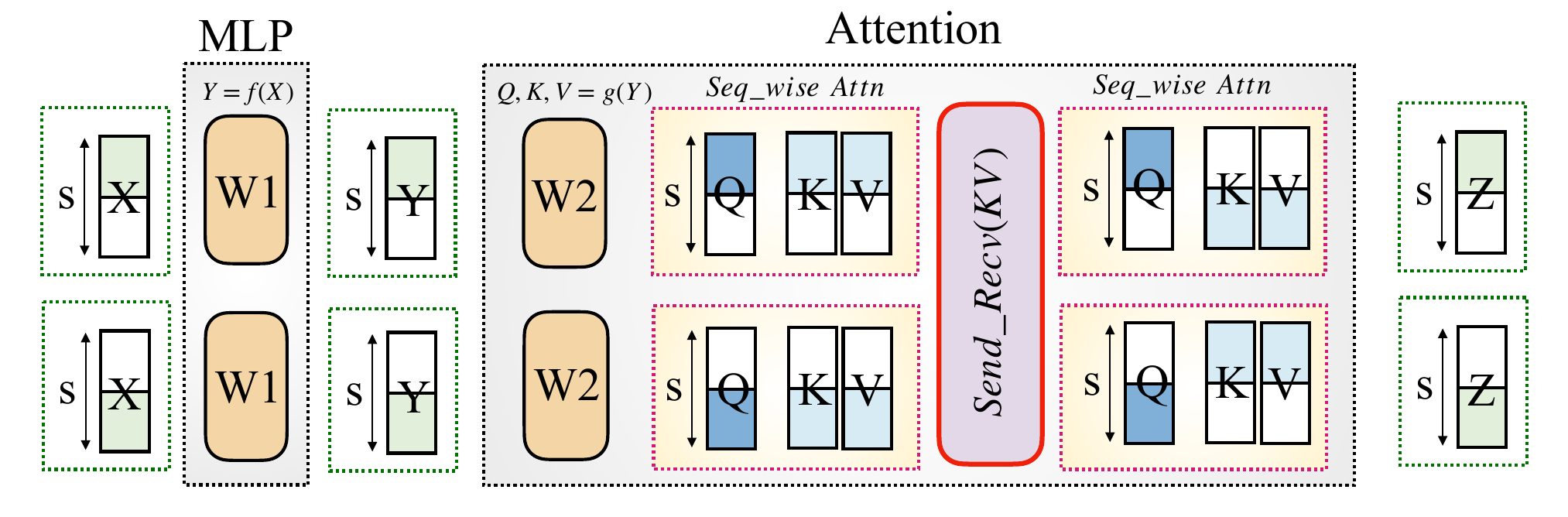}%
\label{fig:headcp}%
}\vfill
\subfloat[Context Parallelism with sequence-wise attention]{%
\includegraphics[width=\linewidth]{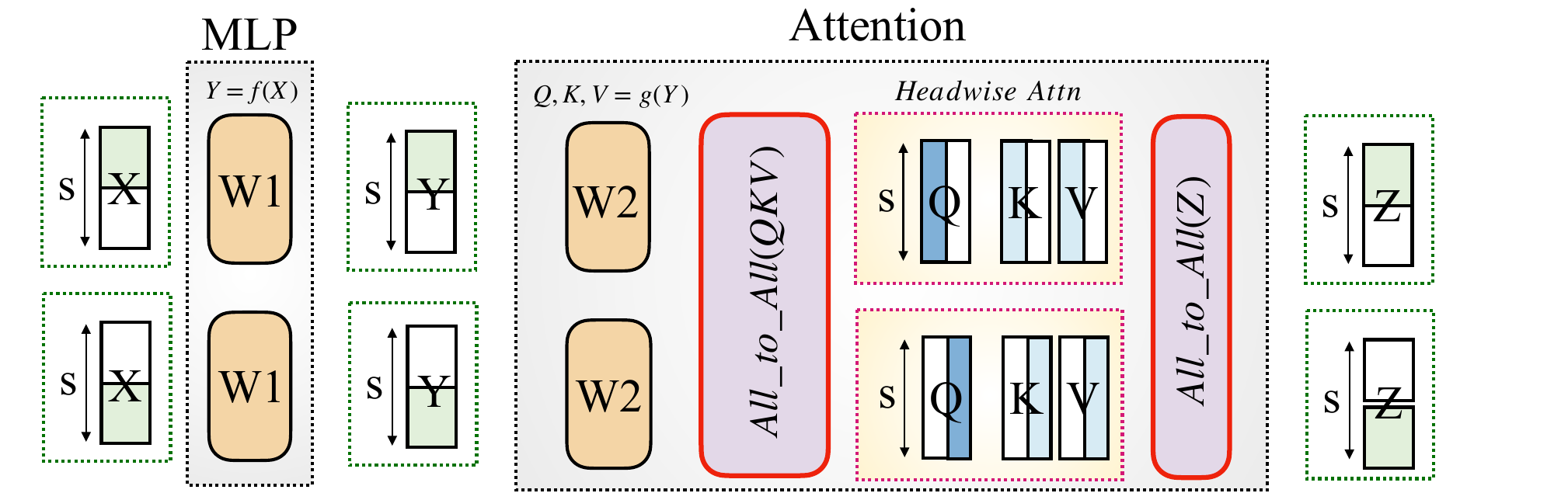}%
\label{fig:ringcp}%
}
\caption{Parallel strategies for DiT inference.}
\label{fig:bg_parallel}
\end{figure}

Not all parallel methods suit DiT inference. Data Parallelism (DP) and Pipeline Parallelism (PP) are ineffective for diffusion generation due to single-sample inference patterns that prevent batch-level parallelization. This leaves two primary strategies for DiT acceleration: 

As shown in Fig.~\ref{fig:bg_parallel}(a), \textbf{Tensor Parallelism} (TP) shards model weights across devices. Linear layers require All-Reduce operations to aggregate partial results, while attention modules need All-Gather communication to gather head-wise results. Though memory-efficient, TP suffers from frequent synchronization overhead.

\textbf{Context Parallelism} (CP) better addresses long-sequence generation by partitioning sequences across devices, eliminating communication in linear layers while only requiring Key-Value (KV) remote access for attention. Two main CP variants have emerged: Head-CP~\cite{ulysses} (Fig.\ref{fig:bg_parallel}(b)) employs all-to-all communication to redistributes tensor layouts, enabling localized head computation. Ring-CP\cite{ringattention} (Fig.\ref{fig:bg_parallel}(c)) maintains sequence-wise partitioning but exchanges KV blocks through ring communication, allowing each GPU to access the full sequence progressively. Recent work like Unified Sequence Parallelism~\cite{usp} provides frameworks to dynamically switch between these strategies based on workload characteristics.

\subsubsection{Diffusion Modules Feature Reuse}

Feature reuse has emerged as a popular optimization technique in diffusion models, leveraging the observation that consecutive denoising steps often produce similar intermediate features. This similarity enables training-free but lossy acceleration through strategic caching and reuse of features from previous steps, effectively bypassing certain computational dependencies. Different approaches target various aspects of the model: Delta-DiT~\cite{deltadit} reuses layer-wise computational features, Pyramid Attention Broadcast (PAB)\cite{pab} focuses on attention output reuse, while TeaCache\cite{teacache} and TaylorSeer\cite{taylorseer} dynamically reuse DiT's overall outputs based on input patterns. However, feature reuse inevitably introduces approximation errors, making the design of reuse strategies crucial for maintaining generation quality.

\subsection{Motivation}

\subsubsection{Limitation: The Scalability-Quality Dilemma}

Recent generation frameworks have pursued higher performance by combining parallel methods with feature reuse strategies. For example, VideoSys~\cite{videosys2024} integrates Dynamic-SP~\cite{dsp} parallelization with PAB~\cite{pab}-based attention output reuse, while SGLang-Diffusion~\cite{sglang} combines Unified-SP~\cite{usp} parallelization with Cache-DiT's TaylorSeer~\cite{taylorseer}-based reuse mechanism. These approaches demonstrate impressive acceleration within single-node environments. However, their performance deteriorates significantly in multi-node scenarios due to limited cross-node communication bandwidth.

This cross-node communication bottleneck creates a fundamental dilemma: to maintain reasonable performance scaling, these frameworks must adopt more aggressive feature reuse strategies to reduce communication overhead. However, such aggressive reuse inevitably leads to substantial quality degradation. Conversely, attempting to preserve generation quality by limiting feature reuse results in poor scaling efficiency due to increased cross-node communication. This inherent trade-off between distributed scalability and generation quality highlights the limitations of treating parallelization and feature reuse as independent optimizations, suggesting the need for a more integrated approach to distributed generation acceleration.

\subsubsection{Analysis: Spatial Locality of Partition Contribution}

\begin{figure}[t]
\centering
\includegraphics[width=\linewidth]{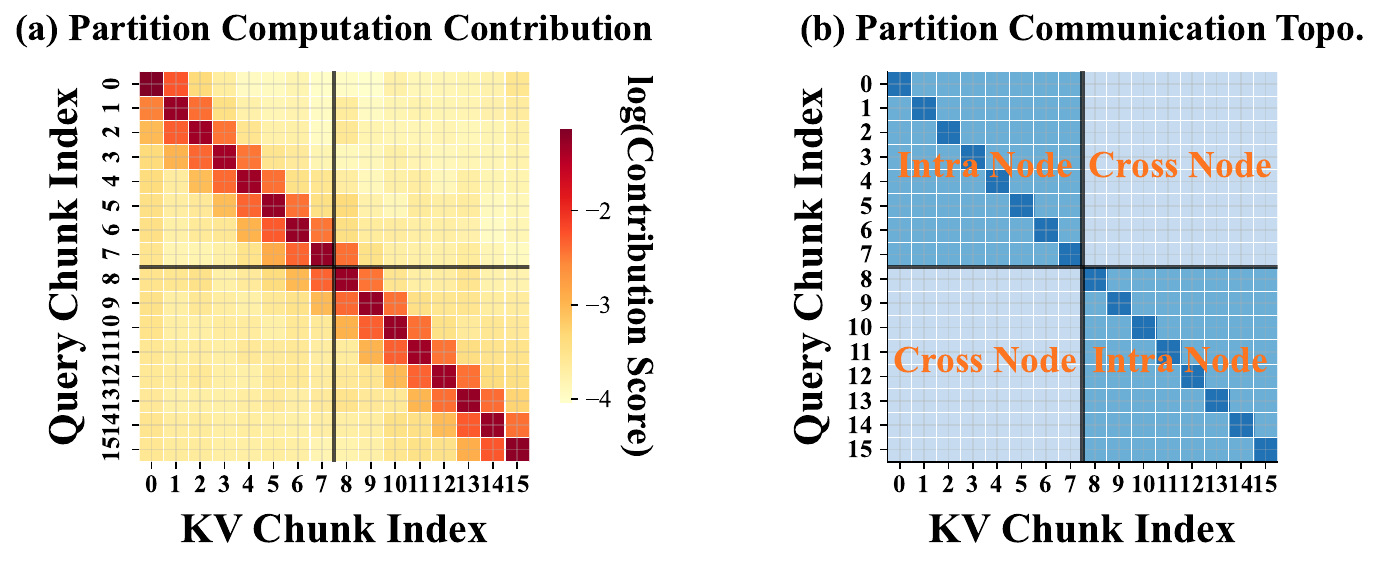}
\vspace{-1em}
\caption{Spatial locality of CP attention in typical cluster: (a) KV partition contribution scores and (b) communication costs across partitions, demonstrating natural alignment between computational importance and system topology.}
\vspace{-1em}
\label{fig:heter}
\end{figure}

To understand this challenge, we analyze the computation contribution of 16 partitions by measuring output differences when omitting specific KV computations, averaged across DiT layers and timesteps using 50 prompts on Wan2.1. Fig.~\ref{fig:heter}(a) reveals a striking pattern of spatial locality: \emph{computational importance strongly correlates with spatial proximity}. Query-Key pairs from nearby partitions contribute significantly more to attention outputs, with importance diminishing as positional distance increases. This spatial locality pattern, consistent across different models and inputs, offers a natural opportunity for optimization when considering system topology.

Theoretically, this pattern demonstrates how fine-grained token-level attention sparsity studied in previous works~\cite{sparsevideogen}, manifests as a simpler distance-dependent pattern when observed at the partition level.

\subsubsection{Insight: System-Pattern Alignment}

This computational pattern naturally aligns with the hierarchical communication topology in distributed systems:
\begin{itemize}[leftmargin=*]
\item \textbf{Local}: KV stored in local memory with negligible access overhead
\item \textbf{Intra-node}: KV from different GPUs within the same node via high-bandwidth NVLink
\item \textbf{Cross-node}: KV from remote nodes via low-bandwidth InfiniBand, incurring substantial cost
\end{itemize}

By mapping spatially adjacent partitions to hardware units with efficient communication, we can achieve perfect alignment between computation importance and communication efficiency. This alignment presents a compelling opportunity: \emph{prioritize high-importance, low-cost computations while replacing expensive operations of minimal contribution with historical results}. This insight enables us to explore previously unreachable optimization space in the scalability-quality trade-off: by prioritizing computations with high contribution-to-cost ratios while reusing historical results for expensive yet less critical operations, we can maximize communication efficiency with minimal quality impact.

A straightforward approach would be caching KV partitions from previous steps. However, this faces a critical \textbf{memory bottleneck}: storing all KV partitions requires $2 \times L \times H \times D \times N \times C$ memory, where $L$=sequence length, $H$=heads, $D$=hidden size, $N$=layers, and $C$=CFG batch size. For Wan-14B ($2 \times 75600 \times 40 \times 128 \times 40 \times 2$ in BF16), this reaches 58GB per GPU, causing OOM when combined with model weights and activations.

\subsubsection{Solution: Attention State Reuse}

To leverage this system-pattern alignment while addressing the memory challenge, we introduce \emph{attention state}, commonly used in partitioned parallel attention computation \cite{bpt,ringattention,flashattn2,flashinfer}, as our optimization medium. For sequence partition $i$, an attention state $\text{AS}_t(i)$ at timestep $t$ comprises output $\mathbf{OUT}_t(i)$ and log-sum-exp $\mathbf{LSE}_t(i)$:

\begin{equation}
\text{AS}_t(i) = \begin{bmatrix}
\mathbf{OUT}_t(i) \\
\mathbf{LSE}_t(i)
\end{bmatrix}, \quad
\begin{aligned}
\mathbf{LSE}_t(i) &= \log \sum_{j \in \mathcal{I}_i} \exp(\mathbf{q}_t \cdot \mathbf{k}_j) \\
\mathbf{OUT}_t(i) &= \sum_{j \in \mathcal{I}_i} \frac{\exp(\mathbf{q}_t \cdot \mathbf{k}_j)}{\exp(\mathbf{LSE}_t(i))} \mathbf{v}_j
\end{aligned}
\label{eq:attention_state}
\end{equation}

where $\mathbf{q}$, $\mathbf{k}$, $\mathbf{v}$ are query, key, and value vectors.

A key property of attention states is their \textbf{composability}—they can be composed associatively and commutatively. For partitions $i$ and $j$:

\begin{equation}
\text{AS}_t(i) \oplus \text{AS}_t(j) = \begin{bmatrix}
\frac{e^{\mathbf{LSE}_t(i)} \mathbf{OUT}_t(i) + e^{\mathbf{LSE}_t(j)} \mathbf{OUT}_t(j)}{e^{\mathbf{LSE}_t(i)} + e^{\mathbf{LSE}_t(j)}} \\
\log(e^{\mathbf{LSE}_t(i)} + e^{\mathbf{LSE}_t(j)})
\end{bmatrix}
\label{eq:composition}
\end{equation}

This composability enables flexible partition-wise computation and reuse strategies.

\begin{figure}[!t]
\centering
\includegraphics[width=0.8\linewidth]{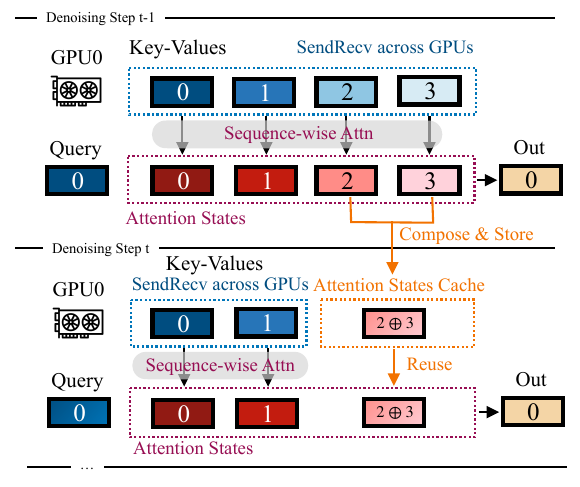}
\caption{Selective attention state reuse mechanism (CP=4, rank=0).}
\label{fig:selective_reuse}
\end{figure}

Leveraging these properties, we design a selective compute and reuse mechanism. Taking CP=4 and rank=0, as an example (Fig.~\ref{fig:selective_reuse}):

In denoising step $t-1$, after computing attention states from partitioned KVs through sequence-wise attention, we strategically compose states from remote partitions ($AS_{t-1}(2)$ and $AS_{t-1}(3)$) and cache the result.

In step $t$, we only compute fresh states for local and near partitions ($AS_{t}(0)$ and $AS_{t}(1)$), while reusing the cached composed state ($AS_{t-1}(2\oplus3)$) for remote partitions, effectively bypassing both KV transmission and attention computation overhead.

This attention state-based approach provides two key advantages over naive KV reuse:(1) Enhanced acceleration by eliminating both communication and computation costs for remote partitions.
(2) Improved memory efficiency through both compact state representation (half of KV size) and flexible state composition.

\section{\sys overview}

\begin{figure}[t]
  \centering
  \includegraphics[width=0.8\linewidth]{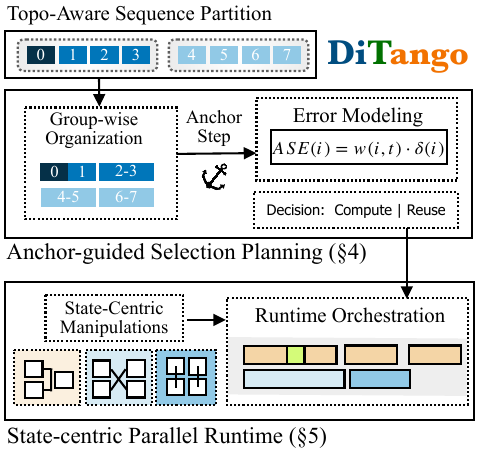}
  \caption{\sys Overview}
  \label{fig:system_overview}
\end{figure}

We propose \sys, a parallel inference system that optimizes diffusion generation through selective attention state reuse. As illustrated in Figure~\ref{fig:system_overview}, \sys has two key components:

\textbf{Anchor-guided selection planning (§\ref{sec:selection}).} Building on error propagation analysis of attention state computation, \sys models the accumulated error introduced by reuse and employs periodic anchor steps for online decision-making. At anchor steps, the system performs fresh computation across all partitions to reset error accumulation, while using the observed attention weight distributions to predict reuse errors for subsequent steps. Through group-wise organization—partitioning the sequence into contiguous groups aligned with device topology—and error budget constraints, the planner generates compute/reuse decisions for each group that balance accuracy and efficiency.

\textbf{State-centric parallel runtime (§\ref{sec:runtime}).} To handle the heterogeneous computation and communication patterns arising from compute/reuse decisions, \sys decouples attention operations into specialized state-centric primitives: Cross-Group Transfer consolidates required KV partitions across devices via symmetric P2P exchanges; Intra-Group Composition computes fresh attention states through ring-based communication within localized device groups; Dynamic Group Compose adaptively merges cached states when memory pressure is detected. The runtime orchestrates these operations across asynchronous computation and communication streams, leveraging dependency-free reuse operations to mask cross-node communication latency and maximize resource utilization throughout the pipeline.

Together, these components enable \sys to exploit spatial locality in diffusion attention while maintaining accuracy guarantees and efficient parallel execution across distributed accelerators.

\section{Anchor-guided Selection Planning}
\label{sec:selection}

\subsection{Error Modeling}

\subsubsection{Attention State Error Propagation}
We first analyze how errors in attention states propagate through composition.
Consider two attention states with errors $\delta\mathbf{OUT}_i$ and $\delta\mathbf{LSE}_i$ for $i \in \{1, 2\}$. Define normalized weights:
\begin{equation}
w_i = \frac{e^{\mathbf{LSE}_i}}{e^{\mathbf{LSE}_1} + e^{\mathbf{LSE}_2}}
\end{equation}

The composition operation propagates errors as:
\begin{equation}
\begin{aligned}
\delta\mathbf{OUT} &= w_1\delta\mathbf{OUT}_1 + w_2\delta\mathbf{OUT}_2 \\&\quad + w_1w_2(\mathbf{OUT}_1 - \mathbf{OUT}_2)(\delta\mathbf{LSE}_1 - \delta\mathbf{LSE}_2) \\
\delta\mathbf{LSE} &= w_1\delta\mathbf{LSE}_1 + w_2\delta\mathbf{LSE}_2
\end{aligned}
\label{eq:error_composition}
\end{equation}

For $n$ partitions, the accumulated output error satisfies:
\begin{equation}
\begin{aligned}
    \|\delta\mathbf{OUT}\|_2 \leq 
    &\sum_{i=1}^n w_i\|\delta\mathbf{OUT}_i\|_2 \\
    &+ \sum_{i<j} w_iw_j\|\mathbf{OUT}_i - \mathbf{OUT}_j\|_2 |\delta\mathbf{LSE}_i - \delta\mathbf{LSE}_j|
\end{aligned}
\label{eq:error_bound}
\end{equation}

The second term is second-order and can be safely ignored: (1) the product $w_iw_j$ is small when one partition dominates, and (2) $|\delta\mathbf{LSE}_i - \delta\mathbf{LSE}_j|$ (difference of errors) is typically much smaller than individual errors. This yields:
\begin{equation}
\|\delta\mathbf{OUT}\|_2 \approx \sum_{i=1}^n w_i\|\delta\mathbf{OUT}_i\|_2
\label{eq:simplified_error_bound}
\end{equation}

\subsubsection{Online Attention State Error Model}

Based on Equation~\ref{eq:simplified_error_bound}, we model the Attention State Error (ASE) of partition $i$'s cached state $\text{AS}_{t_c}(i)$ at current timestep $t$ as:

\begin{equation}
\text{ASE}(i, t, t_c) = w(i, t) \cdot \delta(i, t, t_c)
\label{eq:ase_decomposition}
\end{equation}

where $t_c$ is the cache timestep, $w(i, t)$ captures partition importance, and $\delta(i, t, t_c)$ models temporal error growth over cache age $\tau = t - t_c$.

\paragraph{Partition Importance Weight.}

\begin{figure}
    \centering
    \includegraphics[width=\linewidth]{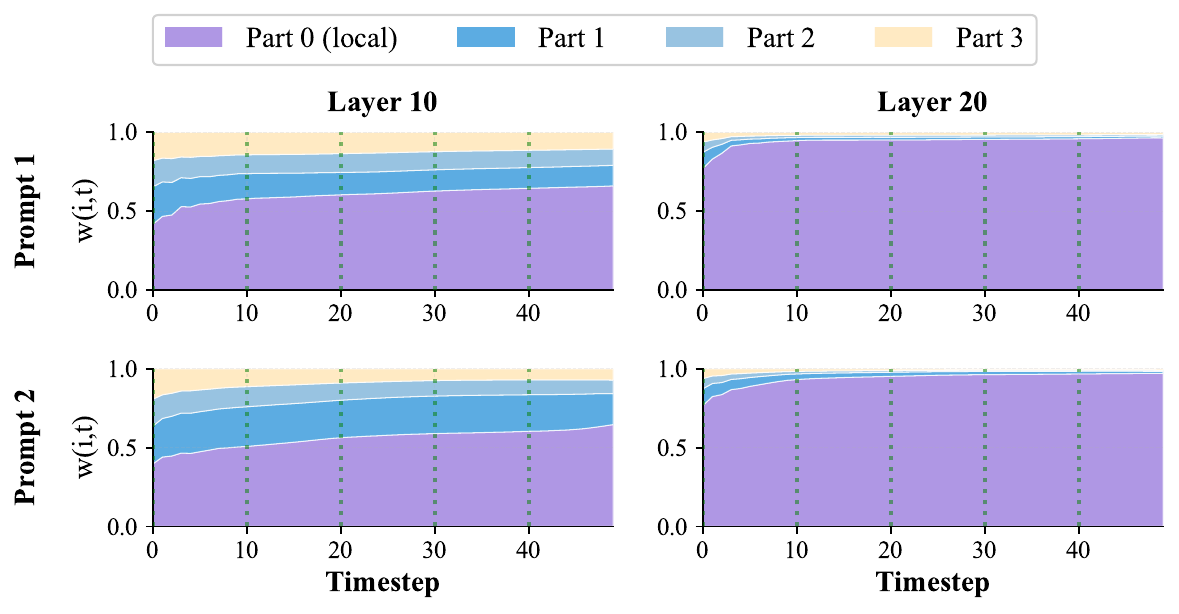}
    \caption{Partition importance $w(i, t)$ with different prompts and layers.}
    \label{fig:lse_weight}
\end{figure}

The weight $w(i, t)$ corresponds to $w_i$ in Equation~\ref{eq:simplified_error_bound}. Since computing exact $w(i, t) = e^{\mathbf{LSE}_t(i)} / \sum_k e^{\mathbf{LSE}_t(k)}$ requires full attention, we leverage the empirical observation that LSE distributions remain stable over $\tau \leq 5$-10 steps (Figure~\ref{eq:lse_weight}). 

We update $w(i, t)$ only at \textbf{anchor steps} $t_a$—timesteps with full attention computation:
\begin{equation}
w(i, t) = w(i, t_a) = \frac{e^{\mathbf{LSE}_{t_a}(i)}}{\sum_{k=1}^n e^{\mathbf{LSE}_{t_a}(k)}}, \quad \forall t \in (t_a, t_a + \tau_{\max}]
\label{eq:lse_weight}
\end{equation}

\paragraph{Age Penalty Error}
Computing $\delta(i, t, t_c) = \|\text{AS}_t(i) - \text{AS}_{t_c}(i)\|_2$ directly contradicts our reuse goal. Instead, we exploit that the \emph{local partition} $i_{\text{loc}}$ (the diagonal partition, always computed) serves as a zero-cost error indicator.

At anchor step $t_a$, we compute the scale ratio:
\begin{equation}
\alpha(i) = \frac{\|\text{AS}_{t_a}(i)\|_2}{\|\text{AS}_{t_a}(i_{\text{loc}})\|_2}
\label{eq:scale_ratio}
\end{equation}

For timesteps $t > t_a$, we extrapolate partition $i$'s error from the local state change:
\begin{equation}
\delta(i, t, t_a) = \alpha(i) \cdot \|\text{AS}_t(i_{\text{loc}}) - \text{AS}_{t_a}(i_{\text{loc}})\|_2
\label{eq:error_extrapolation}
\end{equation}

Our evaluation guarantees that remote states evolve proportionally to the local state, scaled by their relative magnitudes. Since $\text{AS}_t(i_{\text{loc}})$ is computed every step, Equation~\ref{eq:error_extrapolation} incurs nearly zero additional cost while achieving strong empirical accuracy ($R^2 > 0.95$, Section~\ref{sec:accuracy}).

\subsection{Group-wise Compute \& Reuse Selection}

\subsubsection{Group-wise organization}

With a CP size of $p$, KV tensors are distributed across GPUs, each corresponding to an attention stare. To achieve an efficient planning granularity, we group spatially adjacent states. This is motivated by the observed spatial locality that adjacent partitions exhibit similar computational contributions and communication costs, leading to consistent selection decisions. Consequently, the $p$ states are divided into $p/g$ groups, each containing $g$ states. In particular, because the local state is handled separately, the intra-node group containing it has a reduced size of $g-1$.

Organizing partitions as group introduce following benefits:
Grouping strategy naturally reduces cross-term errors in Equation~\ref{eq:error_bound} as spatially adjacent states have similar outputs. It also enables efficient unified scheduling and memory management through group-level attention state composition. 
Building upon this group organization and our error model, we now present the selection strategy that determines which attention state groups to compute or reuse at each timestep.

\subsubsection{Strategy Formulation}

For each timestep $t$, we maintain a set of attention state groups $\mathcal{G} = \{G_0, G_1, ..., G_{m-1}\}$ where $m = \lceil p/g \rceil$. Our goal is to determine for each group $G_i$:

\begin{equation}
\text{Decision}(G_i, t) = \begin{cases}
\texttt{COMPUTE} & \text{compute fresh state} \\
\texttt{REUSE}(t_{c,i}) & \text{reuse cached state from } t_{c,i}
\end{cases}
\end{equation}

where $t_{c,i}$ denotes the last computation timestep for group $G_i$. The strategy balances quality (bounded error) and efficiency (maximized reuse).

\subsubsection{Error-based Selection}

For each group $G_i$ at timestep $t$, we compute its aggregated Attention State Error by summing errors from all partitions within the group:

\begin{equation}
\text{ASE}(G_i, t) = \sum_{j \in G_i} w(j, t) \cdot \delta(j, t, t_{c,i})
\end{equation}

Given an error threshold $\epsilon$, the selection decision follows:

\begin{equation}
\text{Decision}(G_i, t) = \begin{cases}
\texttt{COMPUTE} & \text{if } \text{ASE}(G_i, t) > \epsilon \\
\texttt{REUSE}(t_{c,i}) & \text{otherwise}
\end{cases}
\label{eq:error_based_selection}
\end{equation}

\subsubsection{Anchor Step Enforcement}

To prevent unbounded error accumulation and ensure periodic weight updates (Eq.~\ref{eq:lse_weight} and \ref{eq:scale_ratio}), we enforce \textbf{anchor steps} where all groups perform full computation. An anchor step is triggered when:

\begin{equation}
\text{IsAnchor}(t) = \begin{cases}
\text{True} & \text{if } \forall G_i: \text{ASE}(G_i, t) > \epsilon \\
\text{True} & \text{if } \delta_{\text{local}}(t) > \epsilon_{\text{local}} \\
\text{False} & \text{otherwise}
\end{cases}
\end{equation}

where $\delta_{\text{local}}(t)$ is the relative error of the local state. Unlike prior static thresholds, $\epsilon$ is adapted online as a percentile of the global anchor correction errors to match the target acceleration ratio, while the static $\tau_{\max}$ is replaced by this dynamic local drift condition. At anchor steps, we: (1) force all groups to compute, (2) update partition importance weights $w(j, t)$, (3) update scale ratios $\alpha(j)$, and (4) reset cache timestamps $t_{c,i} \leftarrow t$.

Algorithm~\ref{alg:selection} summarizes our selection strategy. At each timestep, we first compute local attention and predict errors for all groups (lines 2-4). Then we check anchor conditions (lines 5-7). If anchoring is triggered, all groups compute and metadata is updated (lines 8-13); otherwise, groups selectively compute or reuse based on error thresholds (lines 14-21).

\begin{algorithm}[t]
\caption{Attention State Selection Strategy}
\label{alg:selection}
\small
\begin{algorithmic}[1]
\STATE \textbf{Input:} timestep $t$, groups $\mathcal{G}$, cache ratio $R$, local threshold $\epsilon_{\text{local}}$ \STATE Compute $\delta_{\text{local}}(t)$ \FOR{each group $G_i \in \mathcal{G}$}
    \STATE Compute $\text{ASE}(G_i, t) \leftarrow \sum_{j \in G_i} w(j, t) \cdot \delta(j, t, t_{c,i})$ \ENDFOR
\STATE $\text{all\_exceed} \leftarrow \forall G_i: \text{ASE}(G_i, t) > \epsilon$ \STATE $\text{local\_drift} \leftarrow \delta_{\text{local}}(t) > \epsilon_{\text{local}}$ \STATE $\text{is\_anchor} \leftarrow \text{all\_exceed} \lor \text{local\_drift}$ \IF{$\text{is\_anchor}$}
    \FOR{each group $G_i$}
        \STATE $\text{Decision}(G_i, t) \leftarrow \texttt{COMPUTE}$     \ENDFOR
    \STATE Update $\epsilon$ based on $R$ and anchor errors
    \STATE Update $w(j, t)$ and $\alpha(j)$ for all partitions $j$     \STATE $t_{c,i} \leftarrow t$ for all $G_i$ \ELSE
    \FOR{each group $G_i$}
        \IF{$\text{ASE}(G_i, t) > \epsilon$}
            \STATE $\text{Decision}(G_i, t) \leftarrow \texttt{COMPUTE}$, $t_{c,i} \leftarrow t$         \ELSE
            \STATE $\text{Decision}(G_i, t) \leftarrow \texttt{REUSE}(t_{c,i})$         \ENDIF
    \ENDFOR
\ENDIF
\RETURN $\{\text{Decision}(G_i, t)\}$ \end{algorithmic}
\end{algorithm}

\section{State-Centric Attention Runtime}
\label{sec:runtime}

\begin{figure*}[t]
\centering
\includegraphics[width=\textwidth]{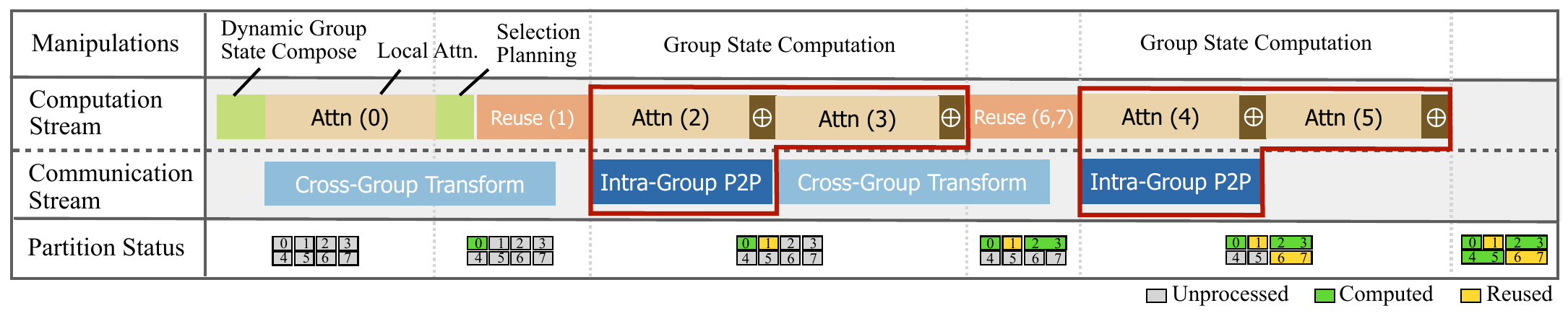}
\caption{Runtime orchestration timeline. The computation and communication streams execute asynchronously, overlapping local attention with Cross-Group Transfer and pipelining multiple Group State Computations.}
\label{fig:runtime_timeline}
\end{figure*}






Given the compute/reuse decisions from selection (Sec.~\ref{sec:selection}), the runtime must efficiently execute heterogeneous partition access patterns while maximizing computation-communication overlap and managing memory constraints. We design a suite of attention state-centric manipulations and orchestrate them in a pipelined runtime system.

\subsection{State-Centric Manipulations}

\begin{figure}[t]
\centering
\includegraphics[width=0.8\linewidth]{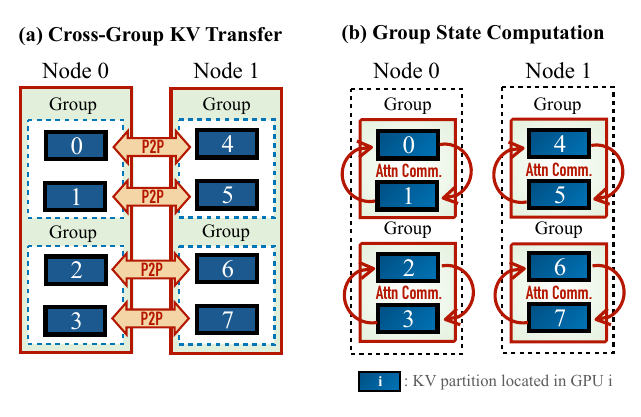}
\caption{State-centric communication operations. (a) Cross-Group KV Transfer: symmetric P2P exchanges reorganize KV partitions across devices to co-locate required data. (b) Group State Computation: intra-group KV circulation enables composed attention state computation within each group.}
\label{fig:group_comm_topo}
\end{figure}

\subsubsection{Cross-Group KV Transfer}

When a group $G_i$ is selected for computation (\texttt{COMPUTE}), we must gather its KV partitions from remote devices to enable local attention computation. At timestep $t$, let $\mathcal{C}_t = {G_i \mid \text{Decision}(G_i, t) = \texttt{COMPUTE}}$ be the set of groups requiring fresh computation. Since our compute/reuse strategy is globally consistent across all devices, each GPU can independently determine its communication peer based on the partition distance within each group.

As illustrated in Figure~\ref{fig:group_comm_topo}(a), the Cross-Group KV Transfer operates through symmetric P2P communication. For each group $G_i \in \mathcal{C}_t$, devices holding partitions of $G_i$ exchange their KV pairs: each GPU performs a \texttt{SendRecv} operation, simultaneously sending its local KV partition to one peer while receiving a different partition from another peer. For example, in a group spanning Node 0 and Node 1, GPU 0 sends its partition to GPU 4 (cross-node) while GPU 4 sends to GPU 0, creating a symmetric exchange pattern.

The goal of this operation is to relocate each required KV partition to the device closest to where it will be consumed, consolidating all partitions of a compute-selected group within a localized device set. By organizing communication at the group granularity, we fully exploit hierarchical bandwidth: intra-node transfers leverage NVLink, while cross-node transfers use InfiniBand, ensuring efficient utilization of available interconnects.

\subsubsection{Group State Computation}
\label{sec:group_state_computation}

Once KV partitions are transferred, each group independently computes its fresh attention state through ring-based composed attention. As shown in Figure~\ref{fig:group_comm_topo}(b), devices within group $G_i$ form a communication ring where KV partitions circulate while attention computation proceeds.

The computation consists of $g$ sequential steps (where $g = |G_i|$ is the group size). At each step, every device: (1) computes partial attention using its current local KV partition and the query, (2) incrementally composes this result with its accumulated attention state using the composition operation from Equation~\ref{eq:error_composition}, and (3) sends its KV partition to the next device in the ring while receiving a new partition from the previous device.

This ring-attention pattern overlaps $(g-1)$ KV transfers with attention computation. The objective of this manipuation is to efficiently produce the complete composed attention state $\text{AS}_t(G_i)$ for group $G_i$ on each participating device. After $g$ steps, every device has processed all partitions in the group and obtained the full composed state, which is then stored in the attention state cache for future reuse. The group-based computation boundary again leverages hierarchical bandwidth, as ring circulation occurs within the localized device set established by the Cross-Group KV Transfer.

\subsubsection{Dynamic Group State Compose}

\sys continuously monitors system memory usage. When memory consumption approaches a predefined threshold $M_{\max}$, the system triggers Dynamic Group State Compose to prevent out-of-memory failures. This manipulation merges existing attention states into coarser-grained groups, exponentially reducing memory overhead.

Given current group size $g$ and a scaling factor $k$ (typically $k=2$), the operation:

\begin{equation}
\begin{aligned}
\text{GroupCompose}(g \rightarrow kg): \quad & AS_t(G_i) \oplus AS_t(G_{i+1}) \oplus \\
... AS_t(G_{i+k-1}) \rightarrow AS_t(G'_{\lfloor i/k \rfloor})
\end{aligned}
\end{equation}

This merging continues recursively until memory usage falls below $M_{\max}$ or an anchor step occurs (which recomputes all states and resets to the original group size $g$). The composed states remain valid for reuse due to the compositional properties of attention states (Eq.~\ref{eq:error_composition}).

\subsection{Runtime Orchestration}

\sys runtime orchestrates the state-centric manipulations across dual asynchronous streams—computation and communication—to maximize overlap while constraining memory usage. Figure~\ref{fig:runtime_timeline} illustrates the pipeline execution flow.

\textbf{Memory-aware initialization.} At each timestep, the runtime first invokes \emph{Dynamic Group State Compose} to monitor memory consumption. If the cached attention states approach the threshold $M_{\max}$, the system merges existing states into coarser groups, exponentially reducing memory overhead. This may trigger re-planning of compute/reuse decisions if the group size $g$ changes, ensuring the selection strategy remains valid under the updated granularity.

\textbf{Local attention and selection planning.} The computation stream then executes local attention on the query's head partition, overlapped with the communication stream's first \emph{Cross-Group KV Transfer}. Concurrently, the runtime performs selection planning for the next timestep, determining which groups to compute or reuse based on the accumulated error budget (Alg.~\ref{alg:selection}). This planning-computation overlap hides the decision latency.

\textbf{Group state computation pipeline.} For each group $G_i \in \mathcal{C}_t$ (the set of groups selected for computation), the runtime sequentially executes:

\begin{enumerate}[leftmargin=*,nosep]
\item Evict stale states: Remove $\text{AS}(G_i)$ from cache, as it will be replaced by the fresh state $\text{AS}_t(G_i)$.
\item \emph{Group State Computation}: Execute ring-based composed attention (Sec.~\ref{sec:group_state_computation}) within the localized device set established by Cross-Group Transfer. The communication stream performs $(g-1)$ intra-group P2P transfers, overlapped with attention computation in the compute stream.
\item Store fresh state: Cache $\text{AS}_t(G_i)$ for future reuse.
\end{enumerate}

This pipeline repeats for all $|\mathcal{C}t|$ compute-selected groups. The runtime dynamically schedules \emph{state reuse} operations—directly loading cached $\text{AS}{t-1}(G_j)$ for reused groups—to fill computation bubbles caused by cross-node communication latency in Cross-Group Transfer. This flexible interleaving ensures high GPU utilization throughout the pipeline.

\textbf{Flexible reuse scheduling.} A key design insight is leveraging \emph{state reuse} operations to mask communication latency. Unlike Fresh Attention State Computation, which depends on prior KV access completing, reuse operations—directly loading cached $\text{AS}_{t-1}(G_j)$ from local memory—have no data dependencies and can execute immediately. The runtime strategically schedules these lightweight reuse operations to fill computation bubbles caused by cross-node communication in Cross-Group Transfer, maintaining high GPU utilization throughout the pipeline.

The "Partition Status" visualization in Figure~\ref{fig:runtime_timeline} illustrates the progress of \sys attention: each partition transitions from unprocessed (gray) to either computed (green) via Group State Computation or reused (yellow) via cached state loading. In the example with 8 partitions, partitions ${0,2,3,4,5}$ are computed while ${1,6,7}$ are reused. Our selection algorithm (Alg.~\ref{alg:selection}) guarantees that by the end of each timestep, all $n$ partitions are covered through either fresh computation or cached reuse, ensuring completeness of the final attention output.

\section{Evaluation}
\sys is implemented in 7K lines of Python (\sysurl), ensuring seamless integration with modern video generation models. We evaluate both its performance and quality improvements, followed by detailed ablation studies to dissect the key factors driving these gains.

\begin{figure*}[t]
    \centering
    \begin{subfigure}{\textwidth}
        \centering
        \includegraphics[width=\textwidth]{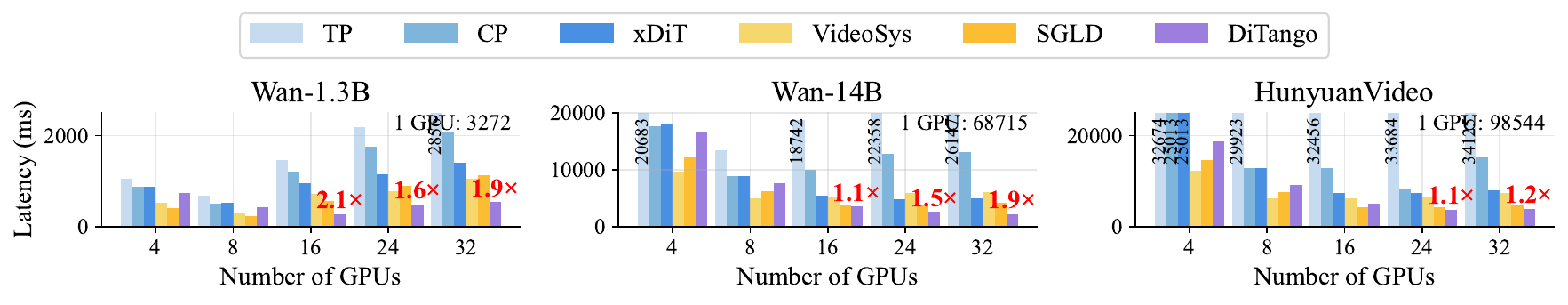}
        \caption{DiT End-to-end latency.}
        \label{fig:eval_e2e}
    \end{subfigure}
    \begin{subfigure}{\textwidth}
        \centering
        \includegraphics[width=\textwidth]{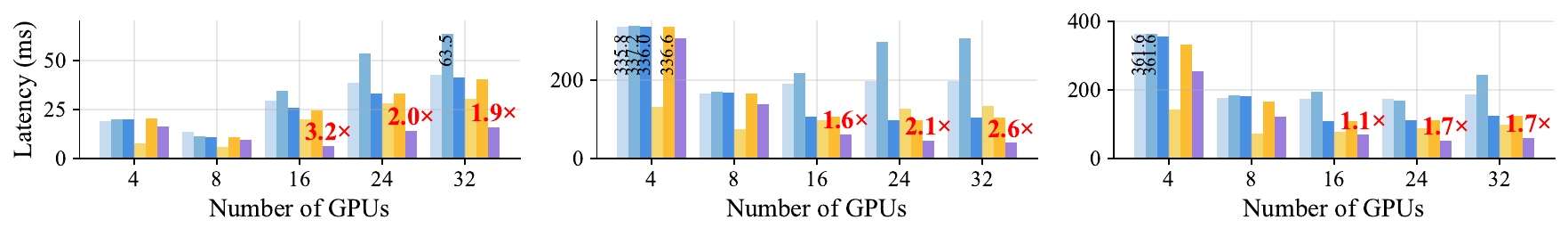}
        \caption{Core Attention latency.}
        \label{fig:eval_attn}
    \end{subfigure}
    \caption{Performance evaluation of \sys. Bars for baselines of too long running time are truncated, and their execution times are marked on the bars. The numbers above \sys's bars show speedups over the best baseline in multi-node inference.}
    \label{fig:performance}
\end{figure*}

\subsection{Evaluation Setup}
\paragraph{Platform}
We evaluate \sys on 4 nodes with 8 NVIDIA H20 GPUs per node (148 TFLOPS FP16, 96GB), with NVLink (900GB/s) for intra-node communication and InfiniBand (400Gbps) for inter-node communication.
Our software stack consists of PyTorch 2.5.0, Python 3.10.16, CUDA 12.4, and GCC 11.4.

\paragraph{Workloads}
We evaluate \sys using the best open-source models on VBench Leaderboard~\cite{vbench}: Wan2.1-14B, Wan2.1-1.3B~\cite{wan}, and HunyuanVideo~\cite{hunyuan}. Table~\ref{tab:model_specs} summarizes their key characteristics. 
The \textit{Tensor Shape} is defined as \textit{[batch size, sequence length, head number, hidden size]}. Wan2.1 models utilize Classifier-Free Guidance\cite{cfg}, with a latent batch size set to 2. For memory-efficient inference, we offload T5 encoder to CPU in Wan2.1-14B. 

\begin{table}[h]
    \centering
    \footnotesize
    \caption{Evaluated Model Specifications}
    \begin{tabular}{c|c|c|c|c|c}
        \hline
        \textbf{Model} & \textbf{Size} & \textbf{Type} & \textbf{Frames} & \textbf{Reso.} & \textbf{Tensor Shape} \\
        \hline
        Wan2.1-14B &  14B & FP16 & 81 & 720×1280 & [2, 76K, 40, 128] \\
        HunyuanVideo & 13B & BF16 & 129 & 720×1280 & [1, 119K, 48, 64] \\
        Wan2.1-1.3B &  1.3B & FP16 & 81 & 832×480 & [2, 33K, 12, 128] \\
        \hline
    \end{tabular}
    \label{tab:model_specs}
\end{table}

\paragraph{Baselines}
We evaluate \sys against five state-of-the-art baselines: Tensor-parallelism~\cite{megatronlm}, the best performing Context Parallel variant between Seq-wise CP~\cite{ringattention} and Head-wise CP~\cite{ulysses}, Hybrid CP~\cite{usp} implemented via xDiT~\cite{xdit}, and two feature reuse-based lossy acceleration systems - VideoSys~\cite{videosys2024,pab} with PAB (Pyramid Attention Broadcast, warm-up/cool-down=4, broadcast range=3) and SGLang-Diffusion~\cite{sglang} (denoted as SGLD) which integrates Cache-DiT~\cite{cache-dit}.

\subsection{DiT End-to-End Performance}
We conduct comprehensive evaluations on both end-to-end DiT inference latency and core attention mechanism performance. In terms of end-to-end performance, \sys demonstrates significant efficiency improvements, achieving up to 1.9× speedup compared to state-of-the-art baselines when performing Wan-14B~\cite{wan} inference across 32 H20 GPUs (Fig.\ref{fig:eval_e2e}). While existing parallel methods such as TP, CP, and xDiT maintain model accuracy through full attention computation, \sys achieves superior performance through selective operation bypassing without compromising generation quality.

In single-node scenarios ($\leq 8$ GPUs), \sys exhibits slightly longer processing times compared to traditional lossy acceleration frameworks VideoSys and SGLang-Diffusion, primarily due to its higher computation ratio designed to preserve generation quality. However, in multi-node deployments, \sys outperforms all baselines by effectively bypassing cross-node communication while maintaining high-quality generation results. This demonstrates \sys's particular effectiveness in distributed computing environments where communication overhead typically becomes a significant bottleneck.

While \sys exhibits strong scalability on Wan-14B and HunyuanVideo, its performance on Wan-1.3B deteriorates beyond 24 GPUs as the limited computation is insufficient to hide intra-node communication latency, resulting in communication-bound behavior.

\subsection{Core Attention Performance \& Efficiency}

\paragraph{Core Attention Latency}
In our evaluation of core attention latency (Fig.\ref{fig:eval_attn}), \sys demonstrates remarkable efficiency, achieving an average 2× speedup compared to all baselines. Notably, this performance advantage extends even over VideoSys~\cite{videosys2024}, despite its aggressive optimization strategy of bypassing approximately one-third of attention computations. SGLang-Diffusion~\cite{sglang} exhibits comparable core attention performance to xDiT~\cite{usp} due to their shared parallel attention implementation, with its optimization is limited to bypassing coarse-grained DiT steps rather than fine-grained attention operations.

\begin{figure}[h]
    \centering
    \includegraphics[width=\linewidth]{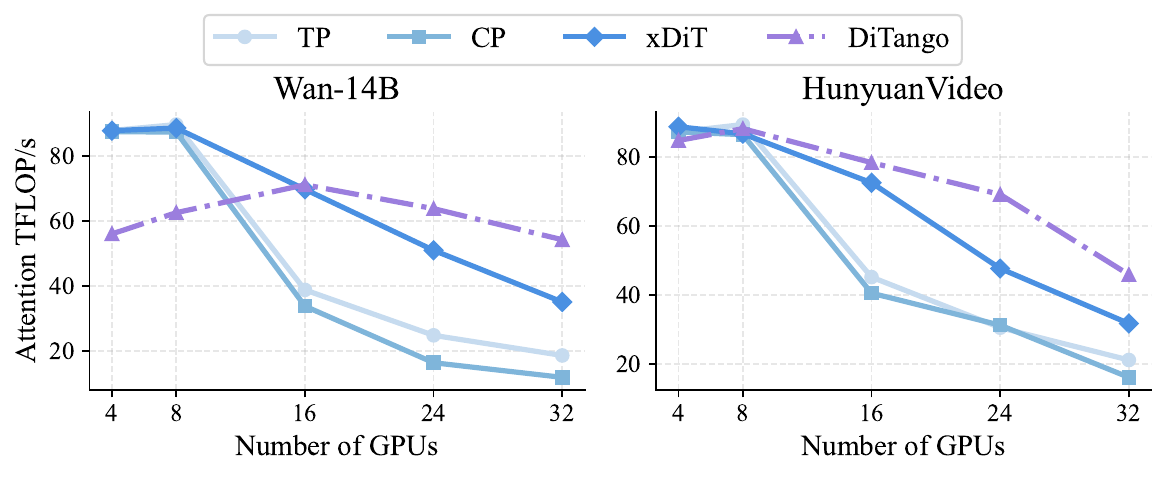}
    \caption{Core attention computational efficiency measured in FLOP/s across different GPU configurations. Note that VideoSys and SGLD is omitted as they utilize the same underlying attention mechanisms with their base implementations (CP and xDiT respectively).}
    \label{fig:eval_tflops}
\end{figure}

\paragraph{FLOP/s}
To further demonstrate the efficiency of \sys attention, we employ FLOP/s metrics to evaluate core attention of different frameworks. Without any kernel-level optimizations, \sys achieves superior computational efficiency in multi-node scenarios through its communication-optimized design, maintaining high FLOP/s where traditional parallel approaches suffer from communication bottlenecks (Fig.~\ref{fig:eval_tflops}). In single-node settings, while \sys shows lower FLOP/s due to extra state-centric operations overhead, its overall latency advantage stems from reduced attention computations compared to traditional approaches that achieve high FLOP/s through computation-communication overlap.



\subsection{Generation Quality \& Tradeoff}

We comprehensively evaluate the generation quality of \sys against existing frameworks using multiple metrics: PSNR, SSIM, and LPIPS for fidelity comparison against original model outputs, and VBench Score~\cite{vbench} for overall generation quality assessment. As shown in Table \ref{tab:quality}, \sys consistently achieves superior performance across most quality metrics while delivering significantly higher speedup across all evaluated models. Notably, baseline methods including TP, CP, and xDiT maintain original model performance through lossless parallelization, thus serving as quality references with optimal VBench Scores but without comparative PSNR, SSIM, or LPIPS measurements.

\begin{table}[t]
    \centering
    \small
    \setlength{\tabcolsep}{4pt}
    \definecolor{lightblue}{RGB}{230,240,255}
    \begin{tabular}{c|c|cccc}
    \toprule
        \multirow{2}{*}{\textbf{Method}} & \textbf{Latency} & \multicolumn{4}{c}{\textbf{Visual Quality}} \\ \cmidrule(lr){2-6}
        ~ & Speedup $\uparrow$ & LPIPS $\downarrow$ & SSIM $\uparrow$ & PSNR $\uparrow$ & VBench(\%) $\uparrow$ \\ 
        \midrule
        \textbf{Wan-1.3B} & 1.20$\times$ & - & - & - & 82.34 \\ \midrule
        VideoSys & 3.13$\times$ & 0.288 & 0.676 & 14.08 & 81.37 \\ 
        SGLD & 2.87$\times$ & 0.220 & 0.728 & 15.77 & 81.92 \\ 
        \rowcolor{lightblue}
        \sys & \textbf{6.08}$\times$ & \textbf{0.214} & \textbf{0.734} & \textbf{16.34} & \textbf{82.21} \\ 
        \midrule
        \textbf{Wan-14B} & 7.02$\times$ & - & - & - & 81.53 \\ \midrule
        VideoSys & 11.25$\times$ & 0.215 & 0.743 & 16.36 & 80.14 \\ 
        SGLD & 16.51$\times$ & 0.261 & 0.707 & 15.02 & 79.86 \\ 
        \rowcolor{lightblue}
        \sys & \textbf{31.15}$\times$ & \textbf{0.137} & \textbf{0.806} & \textbf{18.55} & \textbf{80.79} \\
        \midrule
        \textbf{HunyuanVideo} & 6.26$\times$ & - & - & - & 78.22 \\ \midrule
        VideoSys & 13.28$\times$ & 0.174 & \textbf{0.631} & 21.15 & 77.45 \\ 
        SGLD & 21.00$\times$ & 0.178 & 0.558 & 20.70 & 77.12 \\ 
        \rowcolor{lightblue}
        \sys & \textbf{24.87}$\times$ & \textbf{0.166} & 0.621 & \textbf{21.40} & \textbf{77.89} \\ 
        \bottomrule
    \end{tabular}
    \vspace{1em}
    \caption{Performance and Quality comparison across different models. Speedup ratios compare 32-GPU parallel inference against single-GPU baseline.}
    \label{tab:quality}
\end{table}

\begin{figure}[h]
    \centering
    \includegraphics[width=\linewidth]{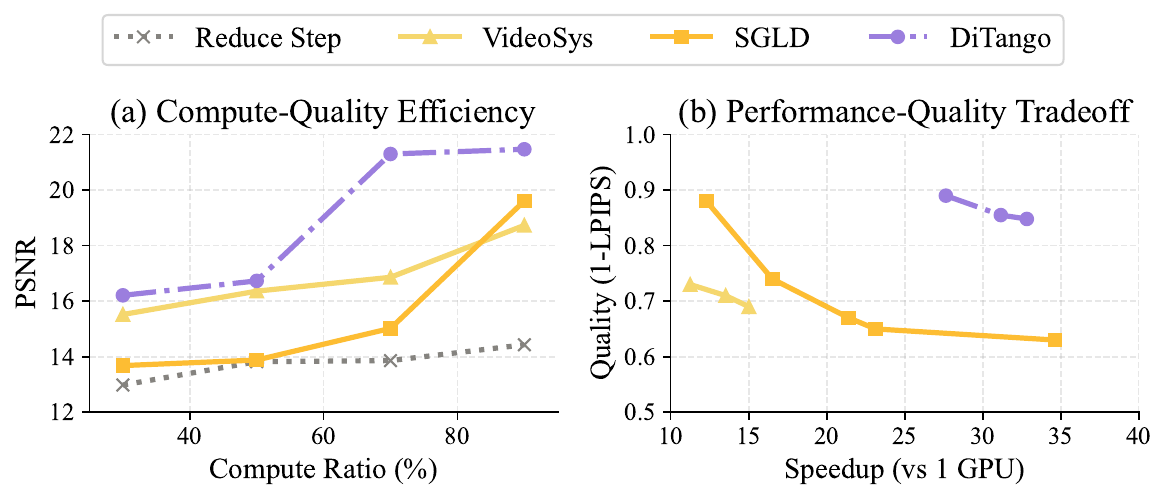}
    \caption{Quality analysis with 32-GPU inference on Wan-14B: (a) Quality comparison under different computation ratios; (b) Quality-speedup tradeoff under different reuse configurations.}
    \label{fig:eval_comp_quality}
\end{figure}

We conduct an in-depth analysis of the factors contributing to \sys's superior generation performance. As illustrated in Fig. \ref{fig:eval_comp_quality}(a), \sys achieves higher generation quality under equivalent computation budgets through two key mechanisms: (1) its contribution-aware approach that precisely identifies and preserves the most crucial computations, and (2) its communication-efficient design that enables more effective computation within given time constraints. These architectural advantages result in a significantly better Pareto frontier in the quality-performance trade-off space, as shown in Fig. \ref{fig:eval_comp_quality}(b), where \sys consistently outperforms existing approaches in balancing generation quality against computational efficiency.

\subsection{Performance Breakdown}

We analyze the effectiveness of key components in \sys.

\paragraph{Error Modeling Accuracy} 

\label{sec:accuracy}

\begin{figure}[t]
    \centering
    \includegraphics[width=0.7\linewidth]{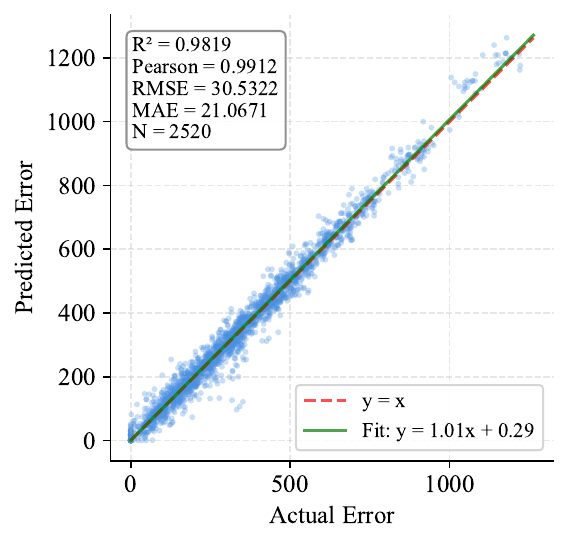}
    \caption{\sys error modeling accuracy.}
    \label{fig:error_accuracy}
\end{figure}

To validate the effectiveness of our error modeling approach, we evaluate the prediction accuracy across diverse samples spanning different timesteps, layers, and prompts. As shown in Fig.~\ref{fig:error_accuracy}, our error model demonstrates remarkable accuracy with an R² value of 0.9792 and a Pearson correlation coefficient of 0.9899, indicating a strong linear relationship between predicted and actual errors. This high correlation validates two key design principles of \sys: (1) the attention state errors across different ranks are indeed highly correlated with local state errors, and (2) our anchor-guided selection mechanism effectively captures error propagation patterns within temporal windows. The low RMSE (32.46) and MAE (21.88) values further confirm that our model can reliably predict error magnitudes, enabling informed decisions in the selection planner for balancing computation efficiency and generation quality.

\paragraph{Runtime Efficiency}

\begin{figure}[h]
\centering
\includegraphics[width=\linewidth]{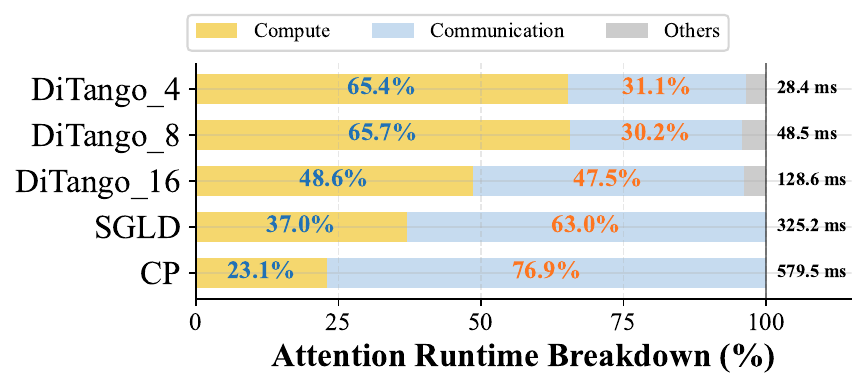}
\caption{Runtime breakdown analysis showing the proportion of computation, communication, and other operations across different approaches and group sizes in 32-GPU distributed inference.}
\label{fig:runtime_efficiency}
\end{figure}

We analyze the runtime efficiency of \sys's state-centric attention mechanism across different state group sizes in a 32-GPU distributed setting. As shown in Fig.~\ref{fig:runtime_efficiency}, \sys consistently achieves higher computation ratios compared to baseline approaches like CP and SGL-Diffusion. This superior computational efficiency is attributed to our effective computation-communication overlap strategy and hierarchical communication design. With group sizes of 4 and 8, \sys demonstrates optimal performance by minimizing cross-node communication. However, at group size 16, while still maintaining better efficiency than baselines, performance is slightly impacted due to necessary inter-group state communications. Notably, the overhead from \sys's additional components (planner execution and memory operations) remains minimal, as indicated by the small "Others" portion in the breakdown.

\paragraph{Extra Overhead}
Despite the complex scheduling logic, \sys introduces minimal system overhead. Computationally, the anchor-guided selection planner incurs negligible overhead. During anchor steps, the additional statistic analysis is performed on norm-compressed representations rather than full-precision tensors, adding less than 1\% to the overall computation compared to a standard full-compute pass. 
On the memory side, maintaining the attention state cache does require additional capacity. For instance, caching the necessary attention states at full precision consumes approximately 30,GB of extra memory for large models like HunyuanVideo and Wan2.1-14B. However, unlike large language models, DiTs possess relatively small parameter footprints. Consequently, this additional memory overhead is well within the capacity of modern GPU architectures. Memory-aware management in section \ref{sec:runtime} is robust enough to ensure practical deployability without out-of-memory constraints.

\subsection{Ablation Study}

\begin{figure}[t]
\centering
\includegraphics[width=\linewidth]{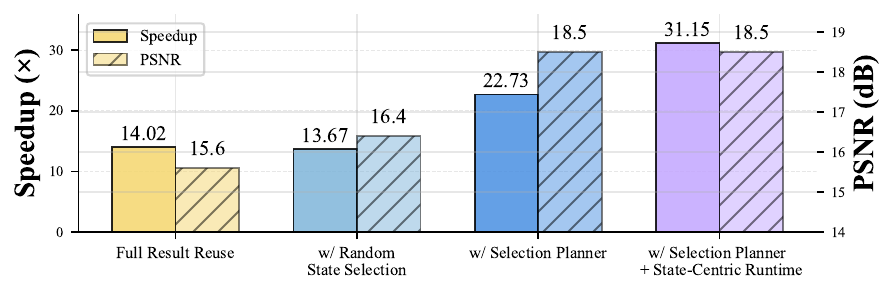}
\caption{Ablation study showing the impact of different \sys components on speedup and generation quality (PSNR) for Wan-14B inference on 32 GPUs.}
\label{fig:ablation}
\end{figure}

We conduct an ablation study to analyze the synergistic effects of \sys's key components on performance and generation quality. As shown in Fig.~\ref{fig:ablation}, using Wan-14B model inference on 32 GPUs as our experimental setting, we evaluate four configurations with increasing system complexity. The baseline approach of full attention result reuse demonstrates the limitations of coarse-grained reuse strategies, showing both modest speedup and generation quality. Introducing fine-grained state reuse with random selection shows limited improvements, as its effectiveness is hampered by unstructured communication patterns and uncontrolled error accumulation. The addition of our selection planner marks a significant improvement in generation quality while also enhancing speedup through more informed state selection. Finally, the complete \sys implementation with both selection planner and state-centric runtime orchestration achieves optimal performance while maintaining high generation quality, demonstrating how these components work in concert to overcome the scalability-quality trade-off.

\section{Related Work}
\label{sec:related work}
\paragraph{Parallelism for Diffusion}
Traditional 3D parallelism strategies like Megatron-LM \cite{megatronlm} exhibit limited efficacy for single-sample diffusion generation. Since the batch size is inherently 1, neither Data Parallelism nor Pipeline Parallelism can be efficiently utilized. While AsyncDiff~\cite{asyncdiff} introduces a timestep-wise pipeline, it fundamentally relies on multi-sample batching, making it inapplicable to single-sample scenarios.

Context Parallelism (CP) \cite{ringattention, ulysses, usp} targets long sequence generation but incurs significant communication overhead when lacking proper state decomposition and topology-aware scheduling. Conversely, Classifier-Free Guidance (CFG) Parallelism~\cite{cfg}, which parallelizes the positive and negative prompt branches, is orthogonal to our approach and can be seamlessly integrated with \sys.

Recently, some approaches have exploited stepwise feature similarity to enhance parallel efficiency. DistriFusion~\cite{distrifusion} leverages asynchronous all-gather for KV cache reuse to mitigate communication in U-Net architectures; however, this design yields sub-optimal results for DiTs due to their differing attention patterns. PipeFusion~\cite{pipefusion, xdit} treats sequence chunks as micro-batches to enable pipeline parallelism, yet it encounters severe consistency degradation, particularly in video generation tasks.

\paragraph{Lossy Diffusion Acceleration}
Diffusion generation is inherently computation-intensive yet robust to noise, making lossy acceleration a mainstream technique for achieving significant speedups with negligible visual degradation. Model-level methods like distillation~\cite{distill1, distill2, recflow} can reduce inference steps by an order of magnitude, but they require extensive extra data and training. At the operator level, quantization~\cite{sageattention1, sageattention2} and sparsity~\cite{sparsevideogen, sparsevideogen2} kernels provide efficient hardware-level acceleration; these techniques are orthogonal to \sys and can be seamlessly integrated with it.

Caching-based methods~\cite{ditfastattn, deltadit, pab, teacache, learningtocache, jano} exploit temporal redundancy across diffusion steps, skipping corresponding computations by reusing similar intermediate features. Compared to distillation, they are training-free and easy to deploy. However, their coarse-grained reuse strategies and reliance on heuristic configurations severely limit their acceleration potential. This limitation becomes particularly pronounced in long video generation, where feature dynamics shift rapidly and require more fine-grained adaptivity. 

\section{Conclusion}

In essence, \sys is a parallel caching framework that strikes an elegant balance between algorithmic design and system orchestration, delivering substantial performance gains without compromising generation quality. It addresses the scalability bottlenecks caused by limited bandwidth in distributed diffusion inference. By selectively reusing attention states, \sys mitigates expensive partition access overhead while preserving critical computations. At its core, the combination of an online selection planner and an efficient runtime engine enables fine-grained compute-reuse control over sequence partitions. Our findings highlight that this granularity is crucial for achieving both system scalability and generation fidelity, which is validated by significant speedups with minimal quality loss over state-of-the-art methods on typical GPU clusters.

Looking ahead, \sys demonstrates strong potential for large-scale generative model serving. It is particularly effective in heterogeneous clusters, where it can reduce generation latency to the second-level, paving the way for real-time and interactive generation applications.

Despite these advancements, we acknowledge two current limitations of \sys. First, in high-bandwidth environments (e.g., single-node setups) where cross-node communication is no longer the primary bottleneck, methods focusing purely on computation reduction may yield higher throughput; thus, a hybrid deployment strategy combining both paradigms is optimal in practice. Second, although \sys incorporates memory-aware management, the attention state cache still consumes substantial capacity, which can constrain GPU utilization in production environments. Future work will focus on addressing this memory footprint through cache compression techniques or intelligent CPU offloading mechanisms to further unleash the potential of parallel diffusion systems.

\begin{acks}
We would like to thank the anonymous reviewers for their insightful comments. This work is partially supported by the Fundamental and Interdisciplinary Disciplines Breakthrough Plan of the Ministry of Education of China (JYB2025XDXM910), NSFC for Distinguished Young Scholar under Grant 62225206, National Natural Science Foundation of China under Grants 62532006, U23A6007, and Beijing Natural Science Foundation under Grant L242017, L243001. Jidong~Zhai is the corresponding author of this paper.
\end{acks}


\bibliographystyle{ACM-Reference-Format}
\bibliography{reference.bib}

@techreport{sora,
  author = {Tim Brooks and Bill Peebles and Connor Holmes and Amos Storkey and Alexei A. Efros and 
            Andreas Terzis and Abhinav Gupta and Devi Parikh and Douwe Kiela and 
            Gabriel Synnaeve and Hannaneh Hajishirzi and Ilya Sutskever and 
            James Zung and Joelle Pineau and Luke Metz and Mira Murati and 
            Pranav Shyam and Rohun Kulkarni and Ruth Fong and Vedant Misra and 
            Yufei Guo and Adrià Recasens and Alexander Papiez and Alexandre Lebrun and 
            Arthur Mensch and Avel Guénin and Bowen Baker and Brandon Houghton and 
            Brian Tanner and Briane Paul V. Samson and Chang Chen and 
            Christopher Clark and Cory McLean and David Martinez-Rubio and 
            David Schnurr and Desu Narendra and Eli Moser and Ethan Perez and 
            Igor Mordatch and J.R. Harris and Joey Faulkner and John Baumgartner and 
            Karl Cobbe and Liam Fedus and Madeleine Thompson and Mark Chen and 
            Mayur Mudigonda and Mimee Xu and Noemi Dreymann and Teddy Lee and 
            Theresa Yoon and Timothy Lillicrap and Torrey Fellbaum and Trevor Darrell and 
            Yuntao Bai and Yutian Chen},
  title = {Video generation models as world simulators},
  institution = {OpenAI},
  year = {2024},
  url = {https://openai.com/research/video-generation-models-as-world-simulators}
}

@misc{sglang,
      title={SGLang: Efficient Execution of Structured Language Model Programs}, 
      author={Lianmin Zheng and Liangsheng Yin and Zhiqiang Xie and Chuyue Sun and Jeff Huang and Cody Hao Yu and Shiyi Cao and Christos Kozyrakis and Ion Stoica and Joseph E. Gonzalez and Clark Barrett and Ying Sheng},
      year={2024},
      eprint={2312.07104},
      archivePrefix={arXiv},
      primaryClass={cs.AI},
      url={https://arxiv.org/abs/2312.07104}, 
}

@article{taylorseer,
  title={From Reusing to Forecasting: Accelerating Diffusion Models with TaylorSeers},
  author={Liu, Jiacheng and Zou, Chang and Lyu, Yuanhuiyi and Chen, Junjie and Zhang, Linfeng},
  journal={arXiv preprint arXiv:2503.06923},
  year={2025}
}

@misc{cache-dit,
  title={cache-dit: A PyTorch-native and Flexible Inference Engine with Hybrid Cache Acceleration and Parallelism for DiTs.},
  url={https://github.com/vipshop/cache-dit.git},
  note={Open-source software available at https://github.com/vipshop/cache-dit.git},
  author={DefTruth, vipshop.com},
  year={2025}
}

@article{wan,
      title={Wan: Open and Advanced Large-Scale Video Generative Models}, 
      author={Team Wan and Ang Wang and Baole Ai and Bin Wen and Chaojie Mao and Chen-Wei Xie and Di Chen and Feiwu Yu and Haiming Zhao and Jianxiao Yang and Jianyuan Zeng and Jiayu Wang and Jingfeng Zhang and Jingren Zhou and Jinkai Wang and Jixuan Chen and Kai Zhu and Kang Zhao and Keyu Yan and Lianghua Huang and Mengyang Feng and Ningyi Zhang and Pandeng Li and Pingyu Wu and Ruihang Chu and Ruili Feng and Shiwei Zhang and Siyang Sun and Tao Fang and Tianxing Wang and Tianyi Gui and Tingyu Weng and Tong Shen and Wei Lin and Wei Wang and Wei Wang and Wenmeng Zhou and Wente Wang and Wenting Shen and Wenyuan Yu and Xianzhong Shi and Xiaoming Huang and Xin Xu and Yan Kou and Yangyu Lv and Yifei Li and Yijing Liu and Yiming Wang and Yingya Zhang and Yitong Huang and Yong Li and You Wu and Yu Liu and Yulin Pan and Yun Zheng and Yuntao Hong and Yupeng Shi and Yutong Feng and Zeyinzi Jiang and Zhen Han and Zhi-Fan Wu and Ziyu Liu},
      journal = {arXiv preprint arXiv:2503.20314},
      year={2025}
}

@inproceedings{
      ditfastattn,
      title={Di{TF}astAttn: Attention Compression for Diffusion Transformer Models},
      author={Zhihang Yuan and Hanling Zhang and Lu Pu and Xuefei Ning and Linfeng Zhang and Tianchen Zhao and Shengen Yan and Guohao Dai and Yu Wang},
      booktitle={The Thirty-eighth Annual Conference on Neural Information Processing Systems},
      year={2024},
      url={https://openreview.net/forum?id=51HQpkQy3t}
}

@article{sparsevideogen,
  author       = {Haocheng Xi and
                  Shuo Yang and
                  Yilong Zhao and
                  Chenfeng Xu and
                  Muyang Li and
                  Xiuyu Li and
                  Yujun Lin and
                  Han Cai and
                  Jintao Zhang and
                  Dacheng Li and
                  Jianfei Chen and
                  Ion Stoica and
                  Kurt Keutzer and
                  Song Han},
  title        = {Sparse VideoGen: Accelerating Video Diffusion Transformers with Spatial-Temporal
                  Sparsity},
  journal      = {CoRR},
  volume       = {abs/2502.01776},
  year         = {2025},
  url          = {https://doi.org/10.48550/arXiv.2502.01776},
  doi          = {10.48550/ARXIV.2502.01776},
  eprinttype    = {arXiv},
  eprint       = {2502.01776},
  bibsource    = {dblp computer science bibliography, https://dblp.org}
}

@article{cogvideox,
  author       = {Zhuoyi Yang and
                  Jiayan Teng and
                  Wendi Zheng and
                  Ming Ding and
                  Shiyu Huang and
                  Jiazheng Xu and
                  Yuanming Yang and
                  Wenyi Hong and
                  Xiaohan Zhang and
                  Guanyu Feng and
                  Da Yin and
                  Xiaotao Gu and
                  Yuxuan Zhang and
                  Weihan Wang and
                  Yean Cheng and
                  Ting Liu and
                  Bin Xu and
                  Yuxiao Dong and
                  Jie Tang},
  title        = {CogVideoX: Text-to-Video Diffusion Models with An Expert Transformer},
  journal      = {CoRR},
  volume       = {abs/2408.06072},
  year         = {2024},
  url          = {https://doi.org/10.48550/arXiv.2408.06072},
  doi          = {10.48550/ARXIV.2408.06072},
  eprinttype    = {arXiv},
  eprint       = {2408.06072},
  bibsource    = {dblp computer science bibliography, https://dblp.org}
}

@misc{kling,
  title = {Kling - Community for short video \& livestream},
  author = {{Kuaishou Technology}},
  year = {2025},
  url = {https://kling.kuaishou.com/en},
  note = {Accessed: 2025-03-07}
}

@article{hunyuan,
  author       = {Weijie Kong and
                  Qi Tian and
                  Zijian Zhang and
                  Rox Min and
                  Zuozhuo Dai and
                  Jin Zhou and
                  Jiangfeng Xiong and
                  Xin Li and
                  Bo Wu and
                  Jianwei Zhang and
                  Kathrina Wu and
                  Qin Lin and
                  Junkun Yuan and
                  Yanxin Long and
                  Aladdin Wang and
                  Andong Wang and
                  Changlin Li and
                  Duojun Huang and
                  Fang Yang and
                  Hao Tan and
                  Hongmei Wang and
                  Jacob Song and
                  Jiawang Bai and
                  Jianbing Wu and
                  Jinbao Xue and
                  Joey Wang and
                  Kai Wang and
                  Mengyang Liu and
                  Pengyu Li and
                  Shuai Li and
                  Weiyan Wang and
                  Wenqing Yu and
                  Xinchi Deng and
                  Yang Li and
                  Yi Chen and
                  Yutao Cui and
                  Yuanbo Peng and
                  Zhentao Yu and
                  Zhiyu He and
                  Zhiyong Xu and
                  Zixiang Zhou and
                  Zunnan Xu and
                  Yangyu Tao and
                  Qinglin Lu and
                  Songtao Liu and
                  Daquan Zhou and
                  Hongfa Wang and
                  Yong Yang and
                  Di Wang and
                  Yuhong Liu and
                  Jie Jiang and
                  Caesar Zhong},
  title        = {HunyuanVideo: {A} Systematic Framework For Large Video Generative
                  Models},
  journal      = {CoRR},
  volume       = {abs/2412.03603},
  year         = {2024},
  url          = {https://doi.org/10.48550/arXiv.2412.03603},
  doi          = {10.48550/ARXIV.2412.03603},
  eprinttype    = {arXiv},
  eprint       = {2412.03603},
  bibsource    = {dblp computer science bibliography, https://dblp.org}
}

@inproceedings{dit,
  author       = {William Peebles and
                  Saining Xie},
  title        = {Scalable Diffusion Models with Transformers},
  booktitle    = {{IEEE/CVF} International Conference on Computer Vision, {ICCV} 2023,
                  Paris, France, October 1-6, 2023},
  pages        = {4172--4182},
  publisher    = {{IEEE}},
  year         = {2023},
  url          = {https://doi.org/10.1109/ICCV51070.2023.00387},
  doi          = {10.1109/ICCV51070.2023.00387},
  bibsource    = {dblp computer science bibliography, https://dblp.org}
}

@article{megatronlm,
  author       = {Mohammad Shoeybi and
                  Mostofa Patwary and
                  Raul Puri and
                  Patrick LeGresley and
                  Jared Casper and
                  Bryan Catanzaro},
  title        = {Megatron-LM: Training Multi-Billion Parameter Language Models Using
                  Model Parallelism},
  journal      = {CoRR},
  volume       = {abs/1909.08053},
  year         = {2019},
  url          = {http://arxiv.org/abs/1909.08053},
  eprinttype    = {arXiv},
  eprint       = {1909.08053},
  bibsource    = {dblp computer science bibliography, https://dblp.org}
}

@article{ringattention,
  author       = {Hao Liu and
                  Matei Zaharia and
                  Pieter Abbeel},
  title        = {Ring Attention with Blockwise Transformers for Near-Infinite Context},
  journal      = {CoRR},
  volume       = {abs/2310.01889},
  year         = {2023},
  url          = {https://doi.org/10.48550/arXiv.2310.01889},
  doi          = {10.48550/ARXIV.2310.01889},
  eprinttype    = {arXiv},
  eprint       = {2310.01889},
  bibsource    = {dblp computer science bibliography, https://dblp.org}
}

@article{ulysses,
  author       = {Sam Ade Jacobs and
                  Masahiro Tanaka and
                  Chengming Zhang and
                  Minjia Zhang and
                  Shuaiwen Leon Song and
                  Samyam Rajbhandari and
                  Yuxiong He},
  title        = {DeepSpeed Ulysses: System Optimizations for Enabling Training of Extreme
                  Long Sequence Transformer Models},
  journal      = {CoRR},
  volume       = {abs/2309.14509},
  year         = {2023},
  url          = {https://doi.org/10.48550/arXiv.2309.14509},
  doi          = {10.48550/ARXIV.2309.14509},
  eprinttype    = {arXiv},
  eprint       = {2309.14509},
  bibsource    = {dblp computer science bibliography, https://dblp.org}
}

@article{dsp,
  author       = {Xuanlei Zhao and
                  Shenggan Cheng and
                  Zangwei Zheng and
                  Zheming Yang and
                  Ziming Liu and
                  Yang You},
  title        = {{DSP:} Dynamic Sequence Parallelism for Multi-Dimensional Transformers},
  journal      = {CoRR},
  volume       = {abs/2403.10266},
  year         = {2024},
  url          = {https://doi.org/10.48550/arXiv.2403.10266},
  doi          = {10.48550/ARXIV.2403.10266},
  eprinttype    = {arXiv},
  eprint       = {2403.10266},
  bibsource    = {dblp computer science bibliography, https://dblp.org}
}

@misc{stepvideo,
      title={Step-Video-T2V Technical Report: The Practice, Challenges, and Future of Video Foundation Model}, 
      author={Guoqing Ma and Haoyang Huang and Kun Yan and Liangyu Chen and Nan Duan and Shengming Yin and Changyi Wan and Ranchen Ming and Xiaoniu Song and Xing Chen and Yu Zhou and Deshan Sun and Deyu Zhou and Jian Zhou and Kaijun Tan and Kang An and Mei Chen and Wei Ji and Qiling Wu and Wen Sun and Xin Han and Yanan Wei and Zheng Ge and Aojie Li and Bin Wang and Bizhu Huang and Bo Wang and Brian Li and Changxing Miao and Chen Xu and Chenfei Wu and Chenguang Yu and Dapeng Shi and Dingyuan Hu and Enle Liu and Gang Yu and Ge Yang and Guanzhe Huang and Gulin Yan and Haiyang Feng and Hao Nie and Haonan Jia and Hanpeng Hu and Hanqi Chen and Haolong Yan and Heng Wang and Hongcheng Guo and Huilin Xiong and Huixin Xiong and Jiahao Gong and Jianchang Wu and Jiaoren Wu and Jie Wu and Jie Yang and Jiashuai Liu and Jiashuo Li and Jingyang Zhang and Junjing Guo and Junzhe Lin and Kaixiang Li and Lei Liu and Lei Xia and Liang Zhao and Liguo Tan and Liwen Huang and Liying Shi and Ming Li and Mingliang Li and Muhua Cheng and Na Wang and Qiaohui Chen and Qinglin He and Qiuyan Liang and Quan Sun and Ran Sun and Rui Wang and Shaoliang Pang and Shiliang Yang and Sitong Liu and Siqi Liu and Shuli Gao and Tiancheng Cao and Tianyu Wang and Weipeng Ming and Wenqing He and Xu Zhao and Xuelin Zhang and Xianfang Zeng and Xiaojia Liu and Xuan Yang and Yaqi Dai and Yanbo Yu and Yang Li and Yineng Deng and Yingming Wang and Yilei Wang and Yuanwei Lu and Yu Chen and Yu Luo and Yuchu Luo and Yuhe Yin and Yuheng Feng and Yuxiang Yang and Zecheng Tang and Zekai Zhang and Zidong Yang and Binxing Jiao and Jiansheng Chen and Jing Li and Shuchang Zhou and Xiangyu Zhang and Xinhao Zhang and Yibo Zhu and Heung-Yeung Shum and Daxin Jiang},
      year={2025},
      eprint={2502.10248},
      archivePrefix={arXiv},
      primaryClass={cs.CV},
      url={https://arxiv.org/abs/2502.10248}, 
}

@misc{pab,
  title={Real-Time Video Generation with Pyramid Attention Broadcast},
  author={Xuanlei Zhao and Xiaolong Jin and Kai Wang and Yang You},
  year={2024},
  eprint={2408.12588},
  archivePrefix={arXiv},
  primaryClass={cs.CV},
  url={https://arxiv.org/abs/2408.12588},
}

@article{xdit,
  author       = {Jiarui Fang and
                  Jinzhe Pan and
                  Xibo Sun and
                  Aoyu Li and
                  Jiannan Wang},
  title        = {xDiT: an Inference Engine for Diffusion Transformers (DiTs) with Massive
                  Parallelism},
  journal      = {CoRR},
  volume       = {abs/2411.01738},
  year         = {2024},
  url          = {https://doi.org/10.48550/arXiv.2411.01738},
  doi          = {10.48550/ARXIV.2411.01738},
  eprinttype    = {arXiv},
  eprint       = {2411.01738},
  bibsource    = {dblp computer science bibliography, https://dblp.org}
}

@inproceedings{bpt,
  author       = {Hao Liu and
                  Pieter Abbeel},
  editor       = {Alice Oh and
                  Tristan Naumann and
                  Amir Globerson and
                  Kate Saenko and
                  Moritz Hardt and
                  Sergey Levine},
  title        = {Blockwise Parallel Transformers for Large Context Models},
  booktitle    = {Advances in Neural Information Processing Systems 36: Annual Conference
                  on Neural Information Processing Systems 2023, NeurIPS 2023, New Orleans,
                  LA, USA, December 10 - 16, 2023},
  year         = {2023},
  url          = {http://papers.nips.cc/paper\_files/paper/2023/hash/1bfd87d2d92f0556819467dc08034f76-Abstract-Conference.html},
  bibsource    = {dblp computer science bibliography, https://dblp.org}
}

@article{flashinfer,
  author       = {Zihao Ye and
                  Lequn Chen and
                  Ruihang Lai and
                  Wuwei Lin and
                  Yineng Zhang and
                  Stephanie Wang and
                  Tianqi Chen and
                  Baris Kasikci and
                  Vinod Grover and
                  Arvind Krishnamurthy and
                  Luis Ceze},
  title        = {FlashInfer: Efficient and Customizable Attention Engine for {LLM}
                  Inference Serving},
  journal      = {CoRR},
  volume       = {abs/2501.01005},
  year         = {2025},
  url          = {https://doi.org/10.48550/arXiv.2501.01005},
  doi          = {10.48550/ARXIV.2501.01005},
  eprinttype    = {arXiv},
  eprint       = {2501.01005},
  bibsource    = {dblp computer science bibliography, https://dblp.org}
}

@inproceedings{flashattn2,
  author       = {Tri Dao},
  title        = {FlashAttention-2: Faster Attention with Better Parallelism and Work
                  Partitioning},
  booktitle    = {The Twelfth International Conference on Learning Representations,
                  {ICLR} 2024, Vienna, Austria, May 7-11, 2024},
  publisher    = {OpenReview.net},
  year         = {2024},
  url          = {https://openreview.net/forum?id=mZn2Xyh9Ec},
  bibsource    = {dblp computer science bibliography, https://dblp.org}
}

@misc{videosys2024,
  author={VideoSys Team},
  title={VideoSys: An Easy and Efficient System for Video Generation},
  year={2024},
  publisher={GitHub},
  url = {https://github.com/NUS-HPC-AI-Lab/VideoSys},
}

@misc{mochi,
      title={Mochi 1},
      author={Genmo Team},
      year={2024},
      publisher = {GitHub},
      journal = {GitHub repository},
      howpublished={\url{https://github.com/genmoai/models}}
}

@inproceedings{distrifusion,
  author       = {Muyang Li and
                  Tianle Cai and
                  Jiaxin Cao and
                  Qinsheng Zhang and
                  Han Cai and
                  Junjie Bai and
                  Yangqing Jia and
                  Kai Li and
                  Song Han},
  title        = {DistriFusion: Distributed Parallel Inference for High-Resolution Diffusion
                  Models},
  booktitle    = {{IEEE/CVF} Conference on Computer Vision and Pattern Recognition,
                  {CVPR} 2024, Seattle, WA, USA, June 16-22, 2024},
  pages        = {7183--7193},
  publisher    = {{IEEE}},
  year         = {2024},
  url          = {https://doi.org/10.1109/CVPR52733.2024.00686},
  doi          = {10.1109/CVPR52733.2024.00686},
  bibsource    = {dblp computer science bibliography, https://dblp.org}
}

@article{pipefusion,
  author       = {Jiannan Wang and
                  Jiarui Fang and
                  Aoyu Li and
                  PengCheng Yang},
  title        = {PipeFusion: Displaced Patch Pipeline Parallelism for Inference of
                  Diffusion Transformer Models},
  journal      = {CoRR},
  volume       = {abs/2405.14430},
  year         = {2024},
  url          = {https://doi.org/10.48550/arXiv.2405.14430},
  doi          = {10.48550/ARXIV.2405.14430},
  eprinttype    = {arXiv},
  eprint       = {2405.14430},
  bibsource    = {dblp computer science bibliography, https://dblp.org}
}

@misc{deltadit,
      title={Delta-DiT: A Training-Free Acceleration Method Tailored for Diffusion Transformers}, 
      author={Pengtao Chen and Mingzhu Shen and Peng Ye and Jianjian Cao and Chongjun Tu and Christos-Savvas Bouganis and Yiren Zhao and Tao Chen},
      year={2024},
      eprint={2406.01125},
      archivePrefix={arXiv},
      primaryClass={cs.CV},
      url={https://arxiv.org/abs/2406.01125}, 
}

@article{cfg,
  author       = {Jonathan Ho and
                  Tim Salimans},
  title        = {Classifier-Free Diffusion Guidance},
  journal      = {CoRR},
  volume       = {abs/2207.12598},
  year         = {2022},
  url          = {https://doi.org/10.48550/arXiv.2207.12598},
  doi          = {10.48550/ARXIV.2207.12598},
  eprinttype    = {arXiv},
  eprint       = {2207.12598},
  bibsource    = {dblp computer science bibliography, https://dblp.org}
}

@article{usp,
  title={A Unified Sequence Parallelism Approach for Long Context Generative AI},
  author={Fang, Jiarui and Zhao, Shangchun},
  journal={arXiv preprint arXiv:2405.07719},
  year={2024}
}

@inproceedings{asyncdiff,
  author       = {Zigeng Chen and
                  Xinyin Ma and
                  Gongfan Fang and
                  Zhenxiong Tan and
                  Xinchao Wang},
  editor       = {Amir Globersons and
                  Lester Mackey and
                  Danielle Belgrave and
                  Angela Fan and
                  Ulrich Paquet and
                  Jakub M. Tomczak and
                  Cheng Zhang},
  title        = {AsyncDiff: Parallelizing Diffusion Models by Asynchronous Denoising},
  booktitle    = {Advances in Neural Information Processing Systems 38: Annual Conference
                  on Neural Information Processing Systems 2024, NeurIPS 2024, Vancouver,
                  BC, Canada, December 10 - 15, 2024},
  year         = {2024},
  url          = {http://papers.nips.cc/paper\_files/paper/2024/hash/ad15848baa3932c0d2deabf0e11d1dcd-Abstract-Conference.html},
  bibsource    = {dblp computer science bibliography, https://dblp.org}
}

@misc{learningtocache,
      title={Learning-to-Cache: Accelerating Diffusion Transformer via Layer Caching}, 
      author={Xinyin Ma and Gongfan Fang and Michael Bi Mi and Xinchao Wang},
      year={2024},
      eprint={2406.01733},
      archivePrefix={arXiv},
      primaryClass={cs.LG}
}

@misc{jano,
      title={Jano: Adaptive Diffusion Generation with Early-stage Convergence Awareness}, 
      author={Yuyang Chen and Linqian Zeng and Yijin ZHou and Hengjie Li and Jidong Zhai},
      year={2026},
      eprint={2603.00519},
      archivePrefix={arXiv},
      primaryClass={cs.CV},
      url={https://arxiv.org/abs/2603.00519}, 
}

@InProceedings{vbench,
     title={{VBench}: Comprehensive Benchmark Suite for Video Generative Models},
     author={Huang, Ziqi and He, Yinan and Yu, Jiashuo and Zhang, Fan and Si, Chenyang and Jiang, Yuming and Zhang, Yuanhan and Wu, Tianxing and Jin, Qingyang and Chanpaisit, Nattapol and Wang, Yaohui and Chen, Xinyuan and Wang, Limin and Lin, Dahua and Qiao, Yu and Liu, Ziwei},
     booktitle={Proceedings of the IEEE/CVF Conference on Computer Vision and Pattern Recognition},
     year={2024}
 }

@article{teacache,
  author       = {Feng Liu and
                  Shiwei Zhang and
                  Xiaofeng Wang and
                  Yujie Wei and
                  Haonan Qiu and
                  Yuzhong Zhao and
                  Yingya Zhang and
                  Qixiang Ye and
                  Fang Wan},
  title        = {Timestep Embedding Tells: It's Time to Cache for Video Diffusion
                  Model},
  journal      = {CoRR},
  volume       = {abs/2411.19108},
  year         = {2024},
  url          = {https://doi.org/10.48550/arXiv.2411.19108},
  doi          = {10.48550/ARXIV.2411.19108},
  eprinttype    = {arXiv},
  eprint       = {2411.19108},
  bibsource    = {dblp computer science bibliography, https://dblp.org}
}

@ARTICLE{distill1,
       author = {{Meng}, Chenlin and {Rombach}, Robin and {Gao}, Ruiqi and {Kingma}, Diederik P. and {Ermon}, Stefano and {Ho}, Jonathan and {Salimans}, Tim},
        title = "{On Distillation of Guided Diffusion Models}",
      journal = {arXiv e-prints},
         year = 2022,
        month = oct,
          eid = {arXiv:2210.03142},
        pages = {arXiv:2210.03142},
          doi = {10.48550/arXiv.2210.03142},
archivePrefix = {arXiv},
       eprint = {2210.03142},
 primaryClass = {cs.CV},
       adsurl = {https://ui.adsabs.harvard.edu/abs/2022arXiv221003142M}
}

@misc{distill2,
      title={Adversarial Diffusion Distillation}, 
      author={Axel Sauer and Dominik Lorenz and Andreas Blattmann and Robin Rombach},
      year={2023},
      eprint={2311.17042},
      archivePrefix={arXiv},
      primaryClass={cs.CV},
      url={https://arxiv.org/abs/2311.17042}, 
}

@misc{recflow,
      title={Rectified Flow: A Marginal Preserving Approach to Optimal Transport}, 
      author={Qiang Liu},
      year={2022},
      eprint={2209.14577},
      archivePrefix={arXiv},
      primaryClass={stat.ML},
      url={https://arxiv.org/abs/2209.14577}, 
}

@inproceedings{sageattention1,
  title={SageAttention: Accurate 8-Bit Attention for Plug-and-play Inference Acceleration}, 
  author={Zhang, Jintao and Wei, Jia and Zhang, Pengle and Zhu, Jun and Chen, Jianfei},
  booktitle={International Conference on Learning Representations (ICLR)},
  year={2025}
}

@inproceedings{sageattention2,
  title={Sageattention2: Efficient attention with thorough outlier smoothing and per-thread int4 quantization},
  author={Zhang, Jintao and Huang, Haofeng and Zhang, Pengle and Wei, Jia and Zhu, Jun and Chen, Jianfei},
  booktitle={International Conference on Machine Learning (ICML)},
  year={2025}
}

@article{sparsevideogen2,
  title={Sparse VideoGen2: Accelerate Video Generation with Sparse Attention via Semantic-Aware Permutation},
  author={Yang, Shuo and Xi, Haocheng and Zhao, Yilong and Li, Muyang and Zhang, Jintao and Cai, Han and Lin, Yujun and Li, Xiuyu and Xu, Chenfeng and Peng, Kelly and others},
  journal={arXiv preprint arXiv:2505.18875},
  year={2025}
}

\end{document}